%% file: AIBench-TR.tex
\documentclass[a4paper, 11pt, oneside]{article} % A4 paper size, default 11pt font size and oneside for equal margins

\usepackage{booktabs} % For formal tables

\usepackage{mathptmx} % This is Times font

\usepackage{fancyhdr}
\usepackage[normalem]{ulem}
\usepackage{microtype}
\usepackage{graphicx}
\usepackage{subfigure}
\usepackage{colortbl}
\definecolor{mygray}{gray}{.88}
\usepackage{multirow}
\usepackage{amsmath}
\usepackage{bm}
\usepackage{epsfig}
\usepackage{authblk}

\usepackage[hyphens]{url}
\usepackage{hyperref}
\usepackage{color}
\usepackage{soul}
\usepackage{bbding}
\usepackage{float}
\definecolor{mygray}{gray}{.88}

\usepackage{cite}
\usepackage{amsmath,amssymb,amsfonts}
\usepackage{algorithmic}
\usepackage{graphicx}
\usepackage{textcomp}
\usepackage{xcolor}

\usepackage{lipsum}
\usepackage{authblk}
\newcommand{\ignore}[1]{}
\usepackage[pass]{geometry}
\usepackage{fancyhdr}
\usepackage[normalem]{ulem}
\usepackage[hyphens]{url}
\usepackage{color}
\usepackage{soul}
\usepackage{graphicx}
\usepackage{multirow}
\usepackage{subfigure}
\usepackage{bbding}
\usepackage{colortbl}
\usepackage{tablefootnote}
\definecolor{mygray}{gray}{.88}
\newcommand{\tabincell}[2]{\begin{tabular}{@{}#1@{}}#2\end{tabular}}
\begin{document} 

%%%%封面内容编辑%%%%
\begin{titlepage} % Suppresses headers and footers on the title page

	\centering % Centre everything on the title page
	
	\scshape % Use small caps for all text on the title page
	
	\vspace*{\baselineskip} % White space at the top of the page
	
	%------------------------------------------------
	%	Title
	%------------------------------------------------
	
	\rule{\textwidth}{1.6pt}\vspace*{-\baselineskip}\vspace*{2pt} % Thick horizontal rule
	\rule{\textwidth}{0.4pt} % Thin horizontal rule
	
	\vspace{0.75\baselineskip} % Whitespace above the title
	
	{\LARGE AIBench Training: \\Balanced Industry-Standard AI Training Benchmarking\\} % Title
	
	\vspace{0.75\baselineskip} % Whitespace below the title
	
	\rule{\textwidth}{0.4pt}\vspace*{-\baselineskip}\vspace{3.2pt} % Thin horizontal rule
	\rule{\textwidth}{1.6pt} % Thick horizontal rule
	
	\vspace{2\baselineskip} % Whitespace after the title block
	
	%------------------------------------------------
	%	Subtitle
	%------------------------------------------------
	
	%Subtitle here % Subtitle or further description
	
	\vspace*{3\baselineskip} % Whitespace under the subtitle
	
	%------------------------------------------------
	%	Editor(s)
	%------------------------------------------------
	
% 	Edited By
    % Authors' contributions
	
 	\vspace{9\baselineskip} % Whitespace before the editors
	This paper has been accepted by the 2021 IEEE International Symposium on Performance Analysis of Systems and Software (ISPASS 2021)
% 	Section~\ref{Introduction} was contributed by Jianfeng Zhan and Wanling Gao. Section~\ref{Methodology} was contributed by Jianfeng Zhan, Wanling Gao, Lei Wang and Fei Tang. Section~\ref{identify} was contributed by Wanling Gao, Fei Tang, Chuanxin Lan, Chunjie Luo, Jiahui Dai, Zheng Cao, Xingwang Xiong, Zihan Jiang, Tianshu Hao, Fanda Fan, Fan Zhang, Yunyou Huang, Jianan Chen, Mengjia Du, Rui Ren, Chen Zheng, Daoyi Zheng, Haoning Tang, Kunlin Zhan, Biao Wang, Defei Kong, Tong Wu, Minghe Yu, Chongkang Tan, Huan Li, Xinhui Tian, Yatao Li, Gang Lu, Junchao Shao, Zhenyu Wang, Xiaoyu Wang, and Hainan Ye. Section~\ref{evaluation} was contributed by Fei Tang, Wanling Gao, Chuanxin Lan, and Xu Wen. Section~\ref{section_conclusion} was contributed by Jianfeng Zhan.
	
% 	{\scshape\Large Fei Tang\\ Wanling Gao\\Jianfeng Zhan\\Chunxin Lan\\Xu Wen\\ Lei Wang\\ Chunjie Luo\\Jiahui Dai\\Zheng Cao\\et al.\\ }
	
	%{\scshape\Large Wanling Gao\\ Fei Tang\\ Lei Wang\\Jianfeng Zhan\\Chunxin Lan\\ Chunjie Luo\\Yunyou Huang\\Jiahui Dai\\Hainan Ye\\Zheng Cao\\Daoyi Zheng\\Haoning Tang\\Kent Zhan\\Biao Wang\\Defei Kong\\Shimin Gong\\Minghe Yu\\Chongkang Tan\\Yabin Huang\\Xinhui Tian\\Yatao Li\\Junchao Shao\\Xiaoyu Wang\\Zhenyu Wang\\ } % Editor list
	
% 	\vspace{0.5\baselineskip} % Whitespace below the editor list

	\vfill % Whitespace between editor names and publisher logo
	
	%------------------------------------------------
	%	Publisher
	%------------------------------------------------
	
	%\plogo % Publisher logo
	%\def\BUlogo{\epsfig{file=ICT.pdf,height=3cm}}
	%\includegraphics[scale=0.135]{ICT.pdf}
	\epsfig{file=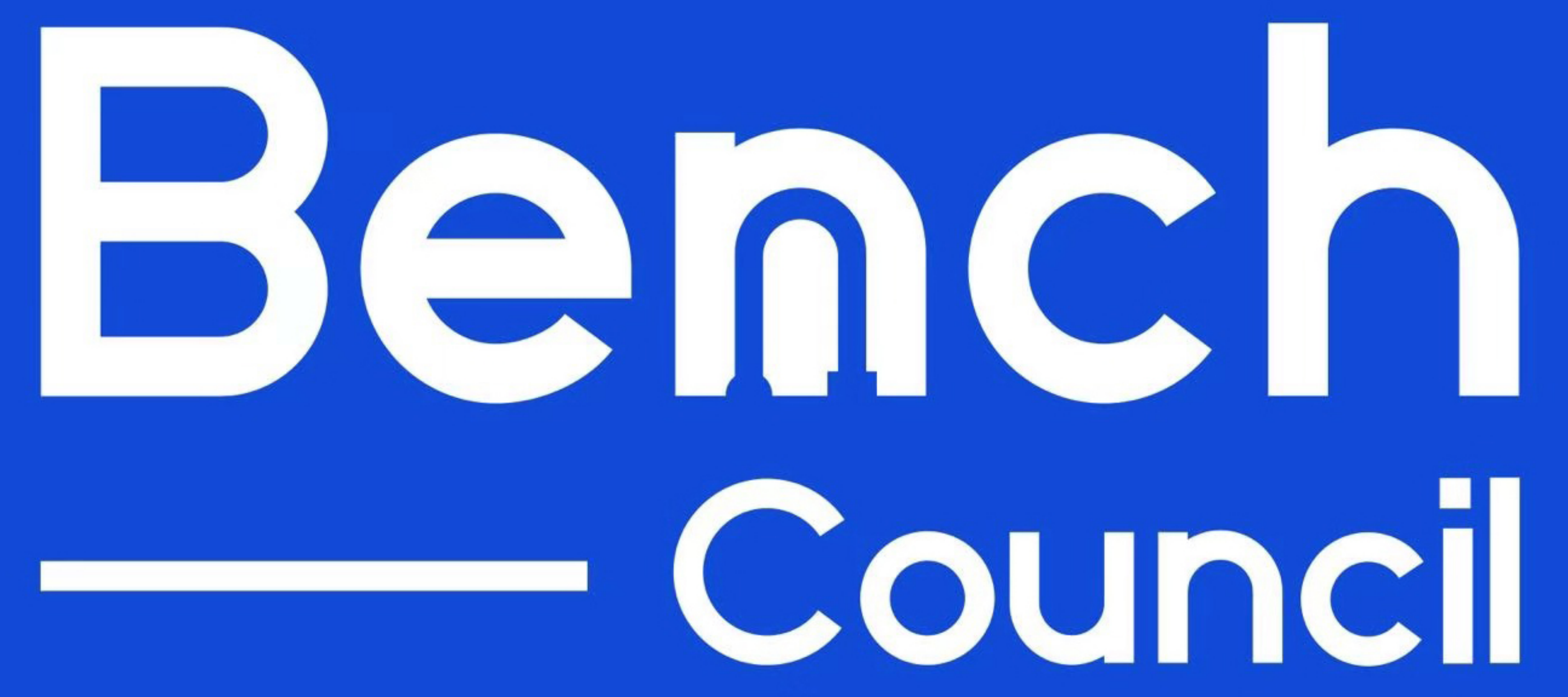,height=2cm}
	\textit{\\BenchCouncil: International Open Benchmarking Council\\Chinese Academy of Sciences\\Beijing, China\\https://www.benchcouncil.org/aibench-training/index.html} % Editor affiliation
	\vspace{1\baselineskip} % Whitespace under the publisher logo
 	
	Technical Report No. BenchCouncil-AIBench-Training-2021-1 % Publication year
	
	{\large March 8, 2020} % Publisher

\end{titlepage}

%----------------------------------------------------------------------------------------

%%%title here%%%
\title{AIBench Training: Balanced Industry-Standard AI Training Benchmarking}

\author[1,3]{Fei Tang}
\author[1,2,3]{Wanling Gao}
\author[1,2,3]{Jianfeng Zhan\thanks{Jianfeng Zhan is the corresponding author.}}
\author[1]{Chuanxin Lan}
\author[1]{Xu Wen}
\author[1,2,3]{Lei Wang}
\author[1,3]{Chunjie Luo}
\author[4]{Zheng Cao}
\author[1]{Xingwang Xiong}
\author[1]{Zihan Jiang}
\author[1]{Tianshu Hao}
\author[1]{Fanda Fan}
\author[1]{Fan Zhang}
\author[2]{Yunyou Huang}
\author[1]{Jianan Chen}
\author[1]{Mengjia Du}
\author[1]{Rui Ren}
\author[1]{Chen Zheng}
\author[5]{Daoyi Zheng}
\author[6]{Haoning Tang}
\author[7]{Kunlin Zhan}
\author[8]{Biao Wang}
\author[9]{Defei Kong}

% \author[21]{Tong Wu}

\author[10]{Minghe Yu}
\author[11]{Chongkang Tan}
\author[12]{Huan Li}
\author[13]{Xinhui Tian}
\author[14]{Yatao Li}
% \author[16]{Gang Lu}
\author[15]{Junchao Shao}
\author[16]{Zhenyu Wang}
\author[17]{Xiaoyu Wang}
\author[2]{Jiahui Dai}
\author[2]{Hainan Ye}

\affil[1]{State Key Laboratory of Computer Architecture, Institute of Computing Technology, Chinese Academy of Sciences \protect\\ \{tangfei, gaowanling, zhanjianfeng, lanchuanxin, wenxu, wanglei\_2011, luochunjie\}@ict.ac.cn}
\affil[2]{International Open Benchmark Council (BenchCouncil)}
%\affil[3]{BenchCouncil R\&D Labs - Beijing, Guilin} 
\affil[3]{University of Chinese Academy of Sciences}
\affil[4]{Alibaba, zhengzhi.cz@alibaba-inc.com}
\affil[5]{Baidu, zhengdaoyi@baidu.com}
\affil[6]{Tencent, haoningtang@tencent.com}
\affil[7]{58.com, zhankunlin@58.com}
\affil[8]{NetEase, bjwangbiao@corp.netease.com}
\affil[9]{ByteDance, kongdefei@bytedance.com}
\affil[10]{Zhihu, yuminghe@zhihu.com}
\affil[11]{Lenovo, tanck1@lenovo.com}
\affil[12]{Paypal, huanli1@paypal.com}
\affil[13]{Moqi, xinhuit@moqi.ai}
\affil[14]{Microsoft Research Asia, yatli@microsoft.com}
% \affil[16]{Huawei, lugang3@huawei.com}
\affil[15]{JD.com, shaojunchao@imdada.cn}
\affil[16]{CloudTa, wangzhenyu@cloudta.com.cn}
\affil[17]{Intellifusion, wang.xiaoyu@intellif.com}
% \affil[21]{China National Institute of Metrology}

\date{March 8, 2021}
\maketitle

\newpage
\begin{abstract}

Earlier-stage evaluations of a new AI architecture/system need affordable AI benchmarks. Only using a few AI component benchmarks like MLPerf alone in the other stages may lead to misleading conclusions. Moreover, the learning dynamics are not well understood, and the benchmarks' shelf-life is short. This paper proposes a balanced benchmarking methodology. 
%We present an AI benchmarking methodology framework to model how learning dynamics are affected by many factors; Then 
We use real-world benchmarks to cover the factors space that impacts the learning dynamics to the most considerable extent. %This methodology can guarantee benchmarks' diversity and representativeness.  
After performing an exhaustive survey on Internet service AI domains, we identify and implement nineteen representative AI tasks with state-of-the-art models.
For repeatable performance ranking (RPR subset) and workload characterization (WC subset), we keep two subsets to a minimum for affordability.   We contribute by far the most comprehensive AI training benchmark suite.
%with seventeen industry partners.

The evaluations show: (1) AIBench Training (v1.1) outperforms MLPerf Training (v0.7) in terms of diversity and representativeness of model complexity, computational cost, convergent rate, computation, and memory access patterns, and hotspot functions; (2) Against the AIBench full benchmarks, its RPR subset shortens the benchmarking cost by 64\%, while maintaining the primary workload characteristics; (3) The performance ranking shows the single-purpose AI accelerator like TPU with the optimized TensorFlow framework performs better than that of GPUs while losing the latter's general support for various AI models. The specification, source code, and performance numbers are available from the AIBench homepage~\url{https://www.benchcouncil.org/aibench-training/index.html}.

\end{abstract}

\include{1_introduction}
\include{2_methodology}
\include{3_representative_ai_tasks}

\include{4_evaluation}

\include{5_performance_ranking}
\include{6_conclusion}
\bibliographystyle{ieeetr}
\bibliography{ref}

\end{document}

%% file: 1_introduction.tex
\section{Introduction}\label{Introduction}

The AI advancements have brought breakthroughs in processing images, video, speech, and audio~\cite{abadi2016tensorflow, smith2017two, hazelwood2018applied},
%Gartner analysts report that AI is oxne of the top ten trends that most impact the infrastructure~\cite{gartner2018}, and predicts that AI will be introduced in almost every products or service by 2020~\cite{gartner2017}. Benefiting from AI technology, many Internet service companies have made significant strides towards improving serving efficiency, 
boosting industry-scale deployments of massive AI algorithms, systems, and architectures. 
%For example, Alibaba proposes a new DUPN network for more effective personalization~\cite{ni2018perceive}. Facebook integrates AI into many essential products and services like news feed~\cite{hazelwood2018applied}. Google proposes the TensorFlow~\cite{abadi2016tensorflow} system and the tensor processing unit (TPU)~\cite{jouppi2017datacenter} to accelerate the service performance. Amazon adopts AI for intelligent product recommendation~\cite{smith2017two}. 
%Consequently, the importance of benchmarking grows. 
%The benchmarks accelerate the process~\cite{hennessy2011computer}, as they provide not only the design inputs but also the evaluation methodology and metrics. %Their relevancy, representativeness, and diversity are of paramount importance as no single benchmark or metric can measure the performance of computer systems on all applications~\cite{gray1993database}.
Unfortunately, the AI training's learning dynamics are dynamic, volatile, and unpredictable: a slight change of models, hyper-parameters, or optimizing strategies may influence the final accuracy or the training convergence rate, which is not understood well theoretically. Meanwhile, the benchmarks' shelf-life is short as state-of-the-art AI models evolve very fast. These situations raise severe AI benchmarking challenges. 
%not well understood theoretically,  Learning dynamics 
% which .
%Unfortunately, there are many factors mutually aggravating the challenges of AI benchmarking.
%in meeting this goal in AI benchmarking. 

 %Our analytics in Section~\ref{identify} shows

\begin{table}[htbp]
\renewcommand\arraystretch{1.2}
\scriptsize
\center
\caption{Comparison of AIBench  Training    and MLPerf Training}\label{comparition_table}
\center
\begin{tabular}{|p{0.5in}|p{1.0in}|p{1.2in}|p{1.2in}|}
% \begin{tabular}{|p{0.38in}|p{0.71in}|p{0.9in}|p{0.87in}|}
%\begin{tabular}{|p{0.9in}|p{0.63in}|p{0.58in}|p{0.55in}|p{0.55in}|p{0.55in}|p{0.4in}|p{0.42in}|}
\hline
 \multicolumn{2}{|c|}{} & \tabincell{c}{AIBench Training v1.1} &\tabincell{c}{MLPerf Training v0.7} \\
%\hline
%\rowcolor{mygray} \multicolumn{9}{|l|}{End-to-End Application Benchmark}\\
%\hline
%\multicolumn{2}{|c|}{Online module} & \CheckmarkBold & $\times$ & $\times$ & $\times$ & $\times$ & $\times$ & $\times$\\
%\hline
%\multicolumn{2}{|c|}{Offline module} & \CheckmarkBold & $\times$ & $\times$ & $\times$ & $\times$ & $\times$ & $\times$\\
\hline
\multicolumn{2}{|l|}{\tabincell{l}{Methodology}} & Balanced methodology considering conflicting requirements & 
According to commercial and research relevance \\
\hline
\multicolumn{2}{|l|}{\tabincell{l}{Task}} & Nineteen tasks and models & six tasks and eight models \\
\hline
\multicolumn{2}{|l|}{\tabincell{l}{Dataset}} & Text, image, 3D, audio, and video data & Text and image data \\
\hline
\multirow{7}{*}{\tabincell{l}{Algorithm\\behavior}} & Computation & 0.09 to  282830 MFLOPs & 0.21 to 29000 MFLOPs \\
\cline{2-4}
& Complexity & 0.03 to 110 million parameters & 5.2 to 110 million parameters \\
\cline{2-4}
& \tabincell{l}{Optimizer categories} & 5 & 5 \\
\cline{2-4}
& Loss function categories & 14 & 6 \\
\cline{2-4}
% & Convergence & 6 to 96 epochs & 3 to 49 epochs \\
\hline
\multirow{1}{*}{\tabincell{l}{System\\behavior}} &
Hotspot functions & 30 & 9\\
\cline{2-4}
& Convergence & 6 to 96 epochs & 3 to 49 epochs \\
\hline
% \multicolumn{2}{|l|}{System behavior} & 30 hot functions & 9 hot functions \\
% \hline
\multirow{6}{*}{\tabincell{l}{Micro-\\architecture\\behavior}} & Achieved occupancy & 0.12 to 0.61 & 0.12 to 0.54 \\
\cline{2-4}
& IPC efficiency & 0.02 to 0.77 & 0.02 to 0.74\\
\cline{2-4}
& Gld efficiency & 0.28 to 0.94 & 0.52 to 0.85\\
\cline{2-4}
& Gst efficiency & 0.27 to 0.98 & 0.75 to 0.98 \\
\cline{2-4}
& DRAM utilization & 0.08 to 0.61 & 0.08 to 0.61 \\
\hline

\end{tabular}
\end{table}

First, the prohibitive cost of training a state-of-the-art AI model raises a serious benchmarking challenge. %even we can cover a full spectrum of AI tasks, models, and data sets, 
 %there are affordability and repeatability challenges. On one hand, 
 %--to.% to achieve the state-of-the-art quality target. 
 Some mixed-precision optimizations indeed improve traditional performance metrics like throughput. Still, they adversely affect the final model's quality, which we can only observe by running an entire training session~\cite{coleman2017dawnbench, mattson2019mlperf}---training an AI model (a component benchmark) to achieve a state-of-the-art quality target.
 So running an entire training session is mandatory. Unfortunately, it is prohibitively costly, often taking several weeks to run a complete training session on a small-scale system. Furthermore, the architecture community heavily relies upon simulations with slowdowns varying wildly from 10X to 1000X, which further exaggerates the challenge.

Second,  there are conflicting benchmarking requirements: affordable vs. comprehensive in different stages.
%, which further aggravates the challenges.
 On the one hand,  earlier-stage evaluations of a new architecture or system need affordable AI benchmarks to reduce the portability cost.  
 Meanwhile, affordable benchmarks are also necessary to provide valuable performance implications in ranking off-the-shelf systems or architectures for promoting its adoption. 
%to promote its adoption,  
 %,while reducing the benchmarking cost.  

On the other hand, later-stage evaluations or purchasing off-the-shelf systems need detailed evaluations using comprehensive benchmarks to avoid benchmarketing~\cite{gray1993database}, and
using a few AI component benchmarks like MLPerf alone 
may lead to misleading or unfair conclusions in the other stages.
For example, 
% using 17 AI component benchmarks of AIBench Training, the training time speedups (only one epoch) on GPU (TITAN V) against that of CPU (Xeon E5-2620 V3) show that not all AI components benefit obviously from the current GPU design. As shown in Fig.~\ref{gpuspeedup}, the speedups vary wildly from 1.78X to 94X, and 3 benchmarks have the lowest speedups, i.e., Spatial Transformer (1.78X), Image Generation (3.5X), and Learning-to-Rank (4.2X).  Meanwhile, 
our experiments find that TPU reflects hugely high performance for Image Classification, 
while supports limited models officially considering the huge portability cost, which is not the case for GPUs. 
Meanwhile, the initial design input or workload characterization needs to consider various computation and memory access patterns to avoid over-optimization for some specific workloads.     %For industry-standard AI benchmarking,  % Fourth, %different stages like the initial design inputs, earlier-stage and later-stage evaluations of a new system/architecture, and  ranking/purchasing commercial off-the-shelf ones have subtly different benchmarking requirements.  The diversity, representativeness, and  relevancy of AI benchmarks are of paramount importance as no single benchmark or metric can measure the performance of computer systems on all applications~\cite{gray1993database}.  
There are other shelf-life, scalability, and repeatability challenges, which we detail in Section~\ref{Challenge}. AIBench is a systematic AI benchmarking project tackling the challenges mentioned above. It distills and abstracts real-world application scenarios across Datacenter, HPC~\cite{jiang_hpc_2021}, IoT~\cite{luo2018iot}, and Edge~\cite{hao2018edge}, into the scenario~\cite{gao2020aibench}, training, inference~\cite{gao_inference_2021}, micro~\cite{tang_micro_2021}, and synthetic benchmarks~\cite{tang_synthetic_2021}, which we detail in Section~\ref{AIBench_Methodology}. This paper focuses on AIBench Training and its subsets.

  We present a balanced methodology to meet conflicting benchmarking requirements in different stages. We use real-world benchmarks to cover the factors space that impacts the learning dynamics to the most considerable extent. The factors which we consider include the commonly used building blocks, model layers, loss functions, optimizers, FLOPs, and different-scale parameter sizes. We identify and include nineteen representative AI tasks from one of the essential domains--Internet Services to guarantee the benchmarks' representativeness and diversity. 
  
  %From the algorithm level, we consider the commonly used building blocks, model layers, loss functions, optimizers, FLOPs, and different-scale parameter sizes; From the system level, we evaluate the convergent rate and hot functions. From the microarchitectural level, we identify their computation and memory access patterns.
  
  We keep two subsets for repeatable performance ranking (RPR subset) and workload characterization (WC subset) to a minimum for affordability.  The criteria for the RPR subset are the diversity of model complexity, computational cost, convergence rate, repeatability, and having widely-accepted metrics or not. The criteria for the WC subset are of micro-architectural characteristics. We currently consider occupancy, IPC, global load, global store, and dram utilization, which are GPU architecture-dependent. 
  %This paper focuses on AIBench training, and another paper, AIBench Inference targeting AI inference, is not discussed here.
  Table~\ref{comparition_table} summarizes the differences of AIBench 
 Training v1.1 against MLPerf Training v0.7~\footnote{We run a smaller dataset for Recommendation  in MLPerf v0.7, and the original dataset is too large to run on a single GPU card.}.

Our contributions are as follows:

\begin{itemize}

%\item We propose and implement a highly extensible, configurable, and flexible AI benchmark framework. %---supporting the construction of multiple end-to-end application benchmarks in a collective way, meanwhile, supporting the evaluation of single component in an individual way.
%The framework provides various AI algorithms targeting at the above problem domains, and is flexible and configurable to constitute a realistic end-to-end application. Meanwhile, each AI algorithm form an individual component or micro benchmark for fine-grained evaluation.

\item We propose a balanced benchmarking methodology that meets conflicting requirements in different stages.

\item We perform by far 
the most comprehensive workload characterization on AIBench (v1.1) and MLPerf (v0.7).
%from the perspectives of model complexity, computational cost, and  convergent rate,  computation and memory access patterns, hotspot functions, and other micro-architecture characteristics. %We reveal that the most important insights.
%From the perspective  of  computation  cost,  the  FLOPs  of  the  AIBenchbenchmarks range from 0.09 to 157802 M-FLOPs, while thatof  MLPerf  vary  from  0.213248  to  24500  M-FLOPs–a  muchnarrower range. From the perspective of model complexity, theamount of learnable parameters of AIBench range from 0.03million  to  68.4  million,  while  MLPerf  only  cover  a  range  of5.2 to 49.53 million. From the perspective of convergent rate,the  required  epochs  range  from  6  to  96,  while  MLPerf only cover a range of 3 to 49.
%\item 
We found that
%AIBench Training covers a much broader range (1.3X to 6.4X) against MLPerf Training in terms of the ratios of peak numbers of model complexity, computational cost, and  convergent rate.
the nineteen benchmarks of AIBench reflect distinct and different computation and memory access patterns from that of MLPerf.
%, emphasizing the necessity of including them for detailed workload characterization and benchmarking.
AIBench outperforms MLPerf in terms of the diversity and representativeness of model complexity, computational cost, convergent rate, computation, memory access patterns, and hotspot functions.  Over MLPerf, AIBench reduces the benchmarking cost while avoiding error-prone design or benchmarketing. 
%\itme Our experiments show that three-dimensional processing like 3D Object Reconstruction has the largest  computational cost, which is about 1.1X greater than image processing on average, and about 129X and 146X greater that that of audio processing and text processing on average, respectively.  While for model complexity, audio processing of Speech Recognition has the largest complexity, which is about 1.2X, 1.4X, and 1.9X larger than that of 3D data, image, and text processing on average, respectively.
%\item We evaluate the benchmark cost of running entire training sessions for all benchmarks in AIBench and MLPerf. The time costs reach to 226 hours for AIBench and more than 362 hours for MLPerf, which is extremely expensive, especially considering the run-to-run variation.

\item We perform a performance ranking of nine state-of-the-practice AI accelerators using the AIBench RPR subset. The ranking list shows the single-purpose AI accelerator like TPU with the optimized TensorFlow framework performs better than that of GPUs while losing the latter's general support for a variety of AI models.

\end{itemize}

The rest of this paper is organized as follows.
% Section \ref{related_work} summarizes the related work.
Section \ref{Methodology} summarizes the challenges, methodology, and related work. 
Section \ref{identify} presents the design and implementation. Section \ref{evaluation} presents the workload characterization. Section \ref{Section_ranking} presents performance ranking.
 Section \ref{section_conclusion} draws a conclusion.

\begin{table*}[htb]
%\scriptsize
\caption{Representative AI Tasks in Internet Service Domain.}
\renewcommand\arraystretch{1.05}
\footnotesize
\label{problemdomain}
\center %p{0.455in}|
\begin{tabular}{|p{0.9in}|p{2.2in}|p{2.2in}|}
\hline
\textbf{Internet Service} & \textbf{Core Scenario} & \textbf{Involved AI Problem Domain} \\
\hline
\multirow{9}{*}{Search Engine} &  \multirow{2}{*}{\parbox{2.2in}{Content-based image retrieval(e.g., face, scene)}} & Object detection; Classification; Spatial transformer; Face embedding; 3D face recognition \\
\cline{2-3}
& Advertising and recommendation &  Advertising; Recommendation \\
\cline{2-3}
& \multirow{2}{*}{Maps search and translation} & 3D object reconstruction; Text-to-Text translation; Speech recognition; Neural architecture search; Nature Language Processing \\
\cline{2-3}
& Data annotation and caption (e.g., text, image) & Text summarization; Image-to-Text \\
\cline{2-3}
& Search result ranking & Learning-to-rank \\
\cline{2-3}
& Image resolution enhancement & Image generation; Image-to-Image \\
\cline{2-3}
& Data storage space and transfer optimization & Image compression; Video prediction \\
\hline

\multirow{9}{*}{Social Network} & Friend or community recommendation & Recommendation; Face embedding; 3D face recognition \\
\cline{2-3}
& Vertical search (e.g., image, people) & Classification; Spatial transformer;  Object detection; Nature Language Processing \\
\cline{2-3}
& Language translation & Text-to-Text translation; Neural architecture search \\
\cline{2-3}
& Automated data annotation and caption & Text summarization; Image-to-Text; Speech recognition \\
\cline{2-3}
& Anomaly detection (e.g., spam image detection) & Classification \\
\cline{2-3}
& Image resolution enhancement & Image generation; Image-to-Image \\
\cline{2-3}
& Photogrammetry (3D scanning) & 3D object reconstruction \\
\cline{2-3}
& Data storage space and transfer optimization & Image compression; Video prediction \\
\cline{2-3}
& News feed ranking & Advertising; Learning-to-rank \\
\hline

\multirow{9}{*}{E-commerce} & Product searching & Classification; Spatial transformer;  Object detection; Nature Language Processing  \\
\cline{2-3}
& Product recommendation and advertising & Recommendation \\
\cline{2-3}
& Language and dialogue translation & Text-to-Text translation; Speech recognition; Neural architecture search \\
\cline{2-3}
& Automated data annotation and caption & Text summarization; Image-to-Text  \\
\cline{2-3}
& Virtual reality (e.g., virtual fitting) & 3D object reconstruction; Image generation; Image-to-Image \\
\cline{2-3}
& Data storage space and transfer optimization & Image compression; Video prediction \\
\cline{2-3}
& Product ranking & Learning to rank \\
\cline{2-3} 
& Facial authentication and payment & Face embedding; 3D face recognition; \\
\hline

\end{tabular}
\end{table*}

%% file: 2_methodology.tex
\section{Challenges, Methodology, and Related Work}~\label{Methodology}

This section presents challenges, methodology, and related work.
%and then details the AIBench Training methodology. Finally, the related work tackling the above challenges is summarized.

%To tackle the above challenges, we take the following methodology.  
% Our balanced AI benchmarking methodology consists of four essential parts as follows.
\subsection{AI Benchmarking Challenges}~\label{Challenge}

Currently, state-of-the-art AI models evolve very fast. The learning dynamics of how huge parameters and system configurations impact are not fully understood, which raises serious AI benchmarking challenges.

First, the prohibitive cost of training a state-of-the-art AI model raises a  serious benchmarking challenge (Cost Challenge \#1). Besides, there are conflicting benchmarking requirements in different stages (Conflicting Requirements Challenge \#2). We detail these two challenges in Section~\ref{Introduction}.

%even we can cover a full spectrum of AI tasks, models, and data sets, 
 %there are affordability and repeatability challenges. On one hand, 
 %--to.% to achieve the state-of-the-art quality target. 
%Some mixed-precision optimizations indeed improve traditional performance metrics like throughput. Still, they adversely affect the final model's quality, which we can only observe by running an entire training session~\cite{coleman2017dawnbench, mattson2019mlperf}---training an AI model (a component benchmark) to achieve a state-of-the-art quality target.
 %So running an entire training session is mandatory. Unfortunately, it is prohibitively costly, often taking several weeks to run a complete training session on a small-scale system. Furthermore, the architecture community heavily relies upon simulations with slowdowns varying wildly from 10X to 1000X, which further exaggerates the challenge. 
   %The component benchmarks are too prohibitively costly to run for simulation-based architecture research. 
  There are massive AI tasks and models. Meanwhile,  an AI model's shelf-life is short~\cite{wang2020systematic} as the state-of-the-art models evolve very fast. 
  Intuitively, to improve affordability, one option is to single out the representative subset from massive AI tasks and models. It is necessary to develop a reasonable AI benchmarking methodology to justify the choices for and update AI tasks, models, and data sets. Another option is to provide different sizes of data inputs. However, a smaller dataset may significantly impact model accuracy, stability, convergence, etc. Another option is to profiling hotspot functions of AI workloads as microbenchmark. The microbenchmarks are affordable to perform a fair comparison of competing systems by isolating hardware and software from statistical  optimizations~\cite{mattson2019mlperf}, but it can not reflect learning dynamics.

Second,  the benchmarks' shelf-life is short~\cite{wang2020systematic} as the state-of-the-art models evolve very fast, which raises another challenge (Shelf-life Challenge \#3).  However, it often takes more than one year to walk through benchmark design, implementation, community adoption, and wide-scale testing. A synthetic benchmark like ParaDNN (2020)~\cite{wang2020systematic} seems useful as it can traverse more expansive parameter spaces. Unfortunately, it can not model learning dynamics: using ParaDNN, we can not observe an entire-training session.  

Third, an AI task's problem scale is often fixed, which raises a scalability challenge (Scalability Challenge \#4). The current benchmarking practices show the largest-scale system can finish an ImageNet training around several ten seconds, which can not constantly stress the system test. Unfortunately, designing a scalable AI benchmark is not trivial. It is essential to model learning dynamics when scaling up a problem size with a synthetic benchmark.
%can not model learning dynamics.  However, benchmarking large-scale systems need to be scalable. 

Finally, the benchmark mandates being repeatable~\cite{jiang_hpc_2021}, while deep learning's nature is stochastic, allowing
multiple different but equally valid solutions~\cite{mattson2019mlperf} (Repeatability Challenge \#5). Previous work manifests HPC AI's uncertainty by run-to-run variation in terms of epochs-to-quality~\cite{jiang_hpc_2021} and the effect of scaling training on time-to-quality~\cite{mattson2019mlperf}.  Run-to-run variation of an AI model depends on many factors, such as the model's initial state, the order of data traversal, and the gradient descent algorithms. The variation between different AI models depends mainly on the model structures.

%Finally, for the same AI task, it is non-trivial to prove the equivalence of two benchmark implementations on different systems or even the same system with different scales (Equivalence Challenge \#6), as we do not fully understand the learning dynamics.  

%Sixth, 
%Scalable means the benchmark can be used to measured different scales of
%system.

%Finally,  promote the wide community use. 

\subsection {AIBench Methodology}~\label{AIBench_Methodology}

\begin{figure}[tb]
\centering
\includegraphics[scale=1.0]{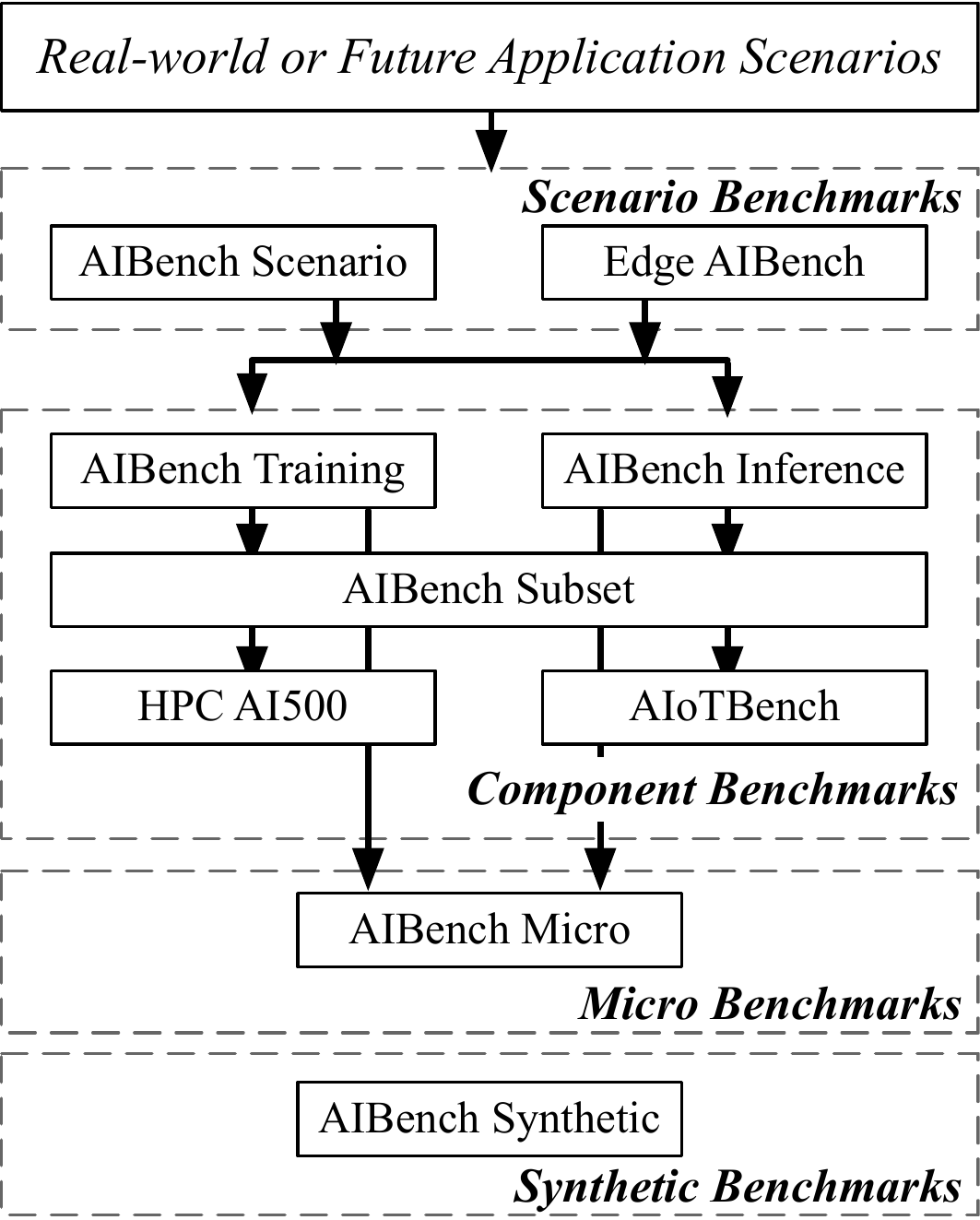}
\caption{AIBench Methodology.} %\vspace{1pt}
\label{aibench-arch}
\end{figure}

%As a joint work with seventeen industry partners, 
%We propose a balanced AI benchmarking methodology to tackle the challenges mentioned above (Challenges \#1-5).
AIBench distills and abstracts real-world application scenarios across Datacenter, HPC (HPC AI500), IoT, and Edge, into the scenario, training, inference, micro, and synthetic benchmarks, as shown in Fig.~\ref{aibench-arch}. 

%for meeting the conflicting requirements among the affordability, representativeness, comprehensiveness and reproducibility in different benchmarking stages. 
%For the later-stage, we propose the comprehensive and representative AI benchmarks (Nineteen  AI tasks). For simulation, we profile the most intensively-used hotspots functions as microbenchmarks that can be used on the simulator. Meanwhile, to improve the affordability, we choose a minimum subset to achieve the affordability. For promoting the wide adoption in the industry community, the benchmarks themselves should be affordable and repeatable, and we choose the subset from the perspective of both architecture-independent and architecture-dependent characteristics, run-to-run variation, and having widely accepted evaluation metrics or not to assure the affordability and repeatability. As a framework, AIBench will be continually updated with new models, dataset, and state-of-the-art quality. %We are adding emerging models like BERT and DLRM, and will update on the website.

AIBench Scenario benchmarks~\cite{gao2020aibench} are proxies to industry-scale real-world applications scenarios. Each scenario benchmark models the critical paths of a real-world application scenario as a permutation of the AI and non-AI modules. Edge AIBench~\cite{hao2018edge} is an instance of the scenario benchmark suites, modeling end-to-end performance across IoT, edge, and Datacenter. 

AIBench Training and AIBench Inference~\cite{gao_inference_2021} cover nineteen representative AI tasks with state-of-the-art models to guarantee diversity and representativeness. AIBench Micro~\cite{tang_micro_2021} provides the intensively-used hotspot functions, profiled from AIBench Training and Inference, for simulation-based architecture researches. AIBench Training 
provides two subsets for repeatable performance ranking (RPR subset) and workload characterization (WC subset) to improve affordability;  It keeps the benchmark subsets to a minimum while maintaining representativeness. 
%provides a subset to improve affordability; 

Based on the AIBench Training RPR subset,  we provide HPC AI500~\cite{jiang2018hpc,jiang_hpc_2021} to evaluate large-scale HPC AI systems. AIoTBench~\cite{luo2018iot} implements AIBench Inference on various IoT and embedded devices, emphasizing diverse light-weight AI frameworks and models. Using a whole-picture workload characterization methodology~\cite{WPC2021TR}, we present the AIBench Inference subset from a perspective of inherent workload characterization.  

As complementary to real-world benchmarks, we plan to provide several synthetic benchmarks (AIBench Synthetic~\cite{tang_synthetic_2021}) with scalable problem sizes to model learning dynamics to tackle Challenges \#3-4.  We aim to achieve scalability, which supports auto-generation of diverse models with different combinations of building blocks and connections, to investigate the impact of varying model structures.
%on system performance and evaluate the upper bound performance of the system under test. In addition, our synthetic workloads contain the accuracy information, which is omitted by the related work.

%\subsubsection{AIBench Training Methodology}
The rest of this subsection details the AIBench Training Methodology. 
AIBench Training adopts a balanced AI benchmarking methodology considering comprehensiveness, representativeness, affordability, and portability. 
Our methodology widely investigates AI tasks and models and covers the algorithm-level, system-level, and microarchitectural-level factors space to the most considerable extent. From the algorithm level, we consider the commonly used building blocks, model layers, loss functions, optimizers, FLOPs, and different-scale parameter sizes; From the system level, we evaluate the convergent rate and hot functions. From the microarchitectural level, we identify their computation and memory access patterns.  We provide real-world AI workloads to achieve comprehensiveness and representativeness. Besides, we propose two AIBench Training subset: RPR and WC subsets to achieve affordability. %for performance ranking and workload characterization.
We give the hotspot functions as microbenchmarks (AIBench Micro) to achieve portability for simulator-based research after profiling. %This paper mainly focuses on the methodology of how to choose and provide real-world AI workloads. 

\subsubsection{Performing a detailed survey of the critical domain rather than a rough survey of a variety of domains}  
As it is impossible to investigate all AI domains, we single out one of the essential AI domains--Internet services for the detailed survey.
In cooperation with seventeen prominent industry partners,  our survey covers diverse business types like Internet services, streaming videos, machine translation, Q\&A community, online payment, etc.

\subsubsection{Include as most as possible representative benchmarks}  
We believe the prohibitive cost of training a model to a state-of-the-art quality cannot justify including only a few AI benchmarks. Instead, using only a few AI component benchmarks may lead to error-prone design: over-optimization for some specific workloads or benchmarketing.  %So AIBench adopts different strategies. We include more diverse benchmarks (16 problem domains, another face forensics is being added) for worload characterization.  

%Meanwhile, we keep a subset of We take two approaches to improving the affordability.
%We choose a minimum (three benchmarks) subset from the  full benchmarks 
%for the proxy of the full benchmark suite.  
%We consider the full benchmarks and the subset as two indispensable parts of the AI benchmark suite, which we call AIBench. 
%Meanwhile, we quantify  the performance relationships among the full benchmarks  and  the subset.
% We run entire training sessions of the subset for the earlier evaluation of a new system or architecture and rank Commercial off-the-shelf systems. 
For Internet services, to the best of our knowledge, we identify and include as most as possible representative AI tasks, models, and datasets into the benchmark suite to guarantee benchmarks' representativeness and diversity. Meanwhile, we consider the primitives' diversity in the AI models. 
% Although the involved primitives of two models may be similar, while their different combinations and execution orders may lead to different behaviors, which is important for workload characterization. 
%Our experiments (Fig.~\ref{time-breakdown}) verify that our benchmarks have different behaviors and time-consuming primitives.
%For each benchmark, we propose the benchmarking rule to assure the fairness across different systems. 
 %We analyze  typical AI application scenarios from three most important Internet services domains, including search engine, social network, and e-commerce, and then we abstract and identify seventeen prominent  AI problem domains, including \emph{ classification, image generation, text-to-text translation, image-to-text, image-to-image, speech-to-text, face embedding, 3D face recognition, object detection, video prediction, image compression, recommendation, 3D object reconstruction, text summarization, spatial transformer, learning to rank, and neural architecture search}.  
 
 %This strategy has two merits: it provides diversity of design inputs for new systems and architectures; it avoids the over-optimization for a few benchmarks.  
 The past successful benchmark practice also witnesses this strategy. The cost of execution time for other benchmarks like HPC~\cite{dongarra2003linpack}, SPECCPU~\cite{speccpu} on simulators, is also prohibitively costly. However, the representativeness and coverage of a widely accepted benchmark suite are paramount important. For example, SPECCPU 2017~\cite{speccpu} contains 43 benchmarks. The other examples include PARSEC3.0 (30)~\cite{bienia2008parsec}, TPC-DS (99)~\cite{tpcds}.
 
\subsubsection{Keep the benchmark subsets to a minimum}%, and explore the performance relationship between the full benchmark suite and its subset} 
%We believe the cost of execution time cannot justify including only a few AI benchmarks. Instead, using only a few AI component benchmarks may lead to error-prone design: over-optimization for some specific workloads.  So AIBench adopts different strategies. We include more diverse benchmarks (16 problem domains, another face forensics is being added) for worload characterization.  
%We take two approaches to improving the affordability.
%On one hand, 
For two different purposes, we choose two minimum  subsets according to different criteria: We choose the RPR subset based on diversity of model complexity, computational cost, convergence rate, repeatability, having the widely-accepted metrics or not, and choose the WC subset according to the representativeness of system or micro-architecture characteristics. % (in short, the subset). 
%for the proxy of the full benchmark suite. 
% Meanwhile, we quantify the performance relationships between the full benchmark suite and its subset.

Using the subset for ranking is also witnessed by the past practice. For example, Top500~\cite{top500}--a super computer ranking--only reports HPL~\cite{dongarra2003linpack} and HPCG~\cite{osti_1361299}--two benchmarks out of 20+ representative HPC benchmarks like HPCC~\cite{luszczek2006hpc}, NPB~\cite{npb}.  

%\subsection{distinguish  from running an  entire training session and a quasi-entire training session}
%Previous work~\cite{zhu2018tbd,gao2018bigdatabench} find that each iteration of an AI task  has the same computation logic  and the iteration number has little impact on micro-architectural behaviors. So we train an AI model to being close to the state-of-the-art  quality target, which we call \emph{a quasi-entire training session}. For an entire training session, we train an AI model to achieve the state-of-the-art  quality target.

\subsubsection{Consider the full benchmarks, their subsets, and microbenchmarks as indispensable}
Different stages have conflicting benchmarking requirements. The initial design inputs to a new system/architecture need comprehensive workload characterization. For earlier-stage evaluations of a new system or architecture, which even adopts simulation-based methods, heavy benchmarking is a significant burden. Thus, concise, portable, and lightweight benchmarks are of great significance. While later-stage evaluations of a new architecture or system or purchasing a commercial off-the-shelf one need detailed evaluations using comprehensive benchmarks to avoid error-prone design or benchmarketing.

%Previous work~\cite{zhu2018tbd,gao2018bigdatabench} find that each iteration of an AI task  has the same computation logic  and the iteration number has little impact on micro-architectural behaviors.  We train the AI models of the seventeen benchmarks to a target quality--that is to run a entire training session.
%, most of which are close to the state-of-the-art ones. We call each training ~\emph{a quasi-entire training session}. 

  %Considering  subtly different benchmarking requirements of different stages, we run entire training  sessions  of  the  subset  and/or selectively run  quasi-entire training  sessions from the full benchmarks to avoid over-optimization  or benchmarketing.  

%\subsection{Consider the comprehensive benchmarks and its subset as two dispensable parts}
For initial design inputs, we perform detailed workload characterization. For later-stage evaluations of or purchasing a new system/architecture, we run the full benchmarks or selectively run some benchmarks to locate the bottlenecks quickly.
% purchasing a new system or architecture
We run an affordable subset for earlier-stage evaluations of a new system/architecture or ranking commercial off-the-shelf systems/architectures. %, as it is portable and lightweight.

\begin{table*}[htb]
\scriptsize
\caption{Component Benchmarks in AIBench.}
\renewcommand\arraystretch{1.4}
\label{AIBench_component}
\center %p{0.455in}|
\begin{tabular}{|p{0.3in}|p{1.0in}|p{1.0in}|p{1.0in}|p{1.0in}|}
\hline
\textbf{No.} & \textbf{Component Benchmark} & \textbf{Algorithm} & \textbf{Dataset} & \textbf{Target Quality}\\
\hline
TrC1 & Image Classification & ResNet50~\cite{he2016deep} &  ImageNet & 74.9\% (accuracy)\\
\hline
TrC2 & Image Generation & WassersteinGAN~\cite{arjovsky2017wasserstein} &  LSUN & N/A \\
\hline
TrC3 & Text-to-Text translation & Transformer~\cite{vaswani2017attention} &  WMT English-German & 55\% (accuracy)  \\
\hline
TrC4 & Image-to-Text & Neural Image Caption Model~\cite{vinyals2017show} &  Microsoft COCO & 4.2 (perplexity)\\
\hline
TrC5 & Image-to-Image Translation & CycleGAN~\cite{zhu2017unpaired} &  Cityscapes & N/A\\
\hline
TrC6 & Speech Recognition & DeepSpeech2~\cite{amodei2016deep} &  Librispeech & 23.5\% (WER)\\
\hline
TrC7 & Face Embedding & Facenet~\cite{schroff2015facenet} &  VGGFace2, LFW & 90\% (accuracy)\\
\hline
TrC8 & 3D Face Recognition & 3D face models~\cite{vieriu2015facial} &  77,715 samples from 253 face IDs & 94.64\% (accuracy) \\
\hline
TrC9 & Object Detection & Faster R-CNN~\cite{ren2015faster} &  VOC2007 & 76\% (mAP)\\
\hline
TrC10 & Recommendation & Neural collaborative filtering~\cite{he2017neural} &  MovieLens & 63.5\% (HR@10) \\
\hline
TrC11 & Video Prediction & Motion-Focused predictive models~\cite{finn2016unsupervised} &  Robot pushing dataset & 72 (MSE)\\
\hline
TrC12 & Image Compression & Recurrent neural network~\cite{toderici2017full} &  ImageNet & 0.99 (MS-SSIM)\\
\hline
TrC13 & 3D Object Reconstruction & Convolutional encoder-decoder network~\cite{yan2016perspective} &  ShapeNet dataset & 45.83\% (IU) \\
\hline
TrC14 & Text Summarization & Sequence-to-sequence model~\cite{nallapati2016abstractive} &  Gigaword dataset & 41 (Rouge-L)\\
\hline
TrC15 & Spatial Transformer & Spatial transformer networks~\cite{jaderberg2015spatial} &  MNIST &  99\% (accuracy)\\
\hline
TrC16 & Learning-to-Rank & Ranking distillation~\cite{tang2018ranking} &  Gowalla & 14\% (accuracy) \\
\hline
TrC17 & Neural Architecture Search & Reinforcement learning~\cite{pham2018efficient} &  PTB~\cite{mitchell1999treebank} & 100 (perplexity)\\
\hline
TrC18 & Advertising & DLRM~\cite{naumov2019deep} & Kaggle Display Advertising Challenge Dataset~\cite{noauthor_kaggle_2014} & 78.9\% (accuracy)\\
\hline
TrC19 & Nature Language Processing (NLP) & BERT~\cite{devlin2018bert} & Wikipedia & 71.2\% (accuracy)\\
\hline
\end{tabular}
\end{table*}

\subsection{Related Work}\label{related_work}

DawnBench (2017)~\cite{coleman2017dawnbench} is the first benchmark that notices the necessity of performing so-called end-to-end benchmarking.  %or running an entire training session. 
It proposes time-to-accuracy as the main metric, which requires training a model to state-of-the-art accuracy.
%The reason is that the optimizations's  adverse effect on the final model's quality can only observe by running an entire training session~\cite{coleman2017dawnbench, mattson2019mlperf}.
AIBench (2018)~\cite{gao2020aibench, gao2018aibench, gao2019aibench} is the first benchmark that notices the necessity of modeling the critical paths of a real-world application scenario. %They noticed When evaluating a new AI model or accelerator with a component benchmark, even it reduces the  tail
%latency significantly, it may contributes little to the overall system tail latency of a real-world application scenario.
ParaDNN (2020)~\cite{wang2020systematic} is the first synthetic AI benchmark. %It uses the combinations of three basic building blocks—fully connected layer, residual/bottleneck block, and RNN/LSTM/GRU cell. The models are only based on the same building blocks in a simple stacking way and cannot reflect the realistic application performance.
The HPL-AI (2019)~\cite{hplai} is a micro benchmark that uses mixed-precision LU decomposition to achieve upper bound FLOPS performance. It is scalable, but LU decomposition is not relevant to most of the AI workloads in AIBench~\cite{jiang_hpc_2021}. Both ParaDNN and HPL-AI can not model the learning dynamics without considering the model quality.

AIBench and MLPerf are two systematic AI benchmarking projects. They are concurrent and complemental.  The  AIBench suites are by far the most comprehensive AI benchmark suites tackling Challenges \#1-5 mentioned above.
MLPerf (2019)~\cite{mattson2019mlperf} includes seven benchmarks for training and five benchmarks for inference. MLPerf performs the most large-scale testing, but it fails to present a benchmarking methodology to justify the choice for and update AI tasks, models, and data sets. Besides, they fail to consider the conflicting requirements, shelf-life, scalability challenges  (Challenges \#2-4) mentioned above.

Other related benchmark projects include DeepBench (2016)~\cite{deepbench},  BenchNN (2012)~\cite{chen2012benchnn}, BenchIP (2018)~\cite{tao2018b}, Fathom (2016)~\cite{adolf2016fathom} and TBD (2018)~\cite{zhu2018tbd}. They either consist of microbenchmarks~\cite{deepbench}, or only evaluate throughput while ignoring model quality\cite{adolf2016fathom, zhu2018tbd} or only include simple shallow neural networks such as multi-layer perceptron~\cite{chen2012benchnn}.

%% file: 3_representative_ai_tasks.tex
\section{The Benchmark Designs and Implementations}\label{identify}

%We first introduction the problem domains identified from industries, including internet service, intelligence medicine, recognition science, etc. Then we illustrate a series of micro and application benchmarks that serve as a service or training model in AIBench framework. 

This section illustrates the benchmark designs and implementations, including selections of workloads, datasets, quality targets (Section~\ref{suiteconst}), metrics, and reference implementations (Section~\ref{rule_metric}). For all benchmarks, we set the target qualities according to the experiments and referenced papers.

\subsection{The Benchmark Decisions}~\label{suiteconst}

To cover a broad spectrum of representative AI tasks in Internet services, we thoroughly analyze the essential application scenarios among three primary Internet services, including search engines, social networks, and e-commerce, as shown in Table~\ref{problemdomain}. In total, we identify nineteen representative AI tasks, for each of which we implement a state-of-the-art model as a component benchmark, as shown in Table~\ref{AIBench_component}. Our benchmarks are constantly evolving and keep updating to use state-of-the-art models. Due to space 
limitation, we give a brief introduction.
%---classification, image generation, text-to-text translation, image-to-text, image-to-image, speech-to-text, face embedding, 3D face recognition, object detection, video prediction, image compression, recommendation, 3D object reconstruction, text summarization, and spatial transformer, and learning to rank.

% \subsubsection{Image Classification}

\textbf{Image Classification} is a fundamental AI task to classify an image into multiple categories, which the industry widely uses.
% many competitions.
% such as the Large Scale Visual Recognition Challenge~\cite{ILSVRC15}.%, which is a supervised learning problem to define a set of target classes and train a model to recognize.

% \textbf{ResNet-50~\cite{he2016deep}}: ResNet-50 is a milestone model which exerts the ability of AI to classify images and exceeds the ability of humans initially. It is a convolutional neural network with 50 layers. %The main module is the bottleneck, consisting of three convolutional layers and a identity mapping. %A milestone in Image Recognition~\cite{he2016deep}, marking the ability of AI to identify images beyond humans. It solves the degradation problem, which means in the very deep neural network, the gradient will gradually disappear in the process of propagation, leading to poor performance. Due to the idea of ResNet, researchers successfully build a 152-layer deep CNN. This ultra deep model won all the awards in ILSVRC'15.

% \textbf{ImageNet Dataset~\cite{deng2009imagenet}}: This dataset is one of the world's largest image database, containing more than 14 million images, and the data size is more than 100 GB. Moreover, it is the most popular dataset in computer vision and is used for many tasks like image classification and object detection.

% \textbf{Reference Quality}: The reference implementation on the ImageNet dataset achieves Top-1 accuracy 74.9\%.

% \subsubsection{Object Detection}
% %Object detection is a fundamental visual recognition problem in computer vision and has been widely studied in the past decades. 
\textbf{Object Detection} aims to find objects of certain target classes with precise localization in each image and
% And like image classification, object detection 
be one of the most important tasks in computer vision.% and assign each object instance a corresponding class label.

\textbf{Learning-to-Rank} is to train models for ranking tasks using machine learning methods. 
\textbf{Image Generation} is to generate similar images with the input image. %Although there are no mature models available for image generation in industry, 

\textbf{Text-to-Text Translation} is to translate a sequence of words into another language and is one of the most representative tasks in natural language processing. %, which is an important task in the field of natural-language processing. It's widely used in daily life, especially in machine translation. In a common text-to-text task, one encoder is used to encode input sequence into embeddings and one decoder to decode embeddings into output sequence. Recurrent neural networks, such as long short-term memory and gated recurrent neural networks, play an important part in those tasks. Convolutional neural networks, which work in encoders and decoders, are uesd as well. While Transfomer\cite{vaswani2017attention}, which was created by Google in 2017, outperforms RNNs and CNNs in both performance and effectiveness.

% \textbf{Transformer~\cite{vaswani2017attention} }: Transformer is the classical model for text translation and is the basis for the subsequent Bert~\cite{devlin2018bert} model. It is combined with self-attention and Feed Forward Neural Network.
% %The encoder is combined of six layers, each layer contains one multi-head self-attention mechanism and one connected feed-forward network. The decoder is combined of six layers, each layer contains two multi-head self-attention mechanism and one connected feed-forward network. 
% %It has a better parallelism than RNN as well as a better performance on accuracy.
% %We use Transformer to translate English into German.

% \textbf{WMT English-German Dataset~\cite{wmt}}: The training dataset is the WMT'14 English-German data, which has 4.5 million sentence pairs. %The inference data set is  newstest2014, which has 2737 sentence pairs.

% \textbf{Reference Quality}: The target accuracy is 55\%. %which is created especially for evaluating the quality of machine-translated texts, as the metric and the reference result is 28.4.

% \subsubsection{Image-to-Text}

% % 1. Task:
\textbf{Image-to-Text} is to generate a description for each image automatically. 
\textbf{Image-to-Image Translation} is to learn image mapping of two different domains, which we use for style conversion, object conversion, seasonal conversion, and photo enhancement.

\textbf{Speech Recognition} 
% is the most important task in audio processing, and it 
is to recognize speech audio and translate it into a text.

% \textbf{DeepSpeech2 Model~\cite{amodei2016deep}}:  DeepSpeech2 is a milestone model in speech recognition. And the model is a recurrent neural network (RNN) with one or more convolutional input layers, followed by multiple recurrent layers and one fully connected layer before a softmax layer~\cite{amodei2016deep}. %For training, Deep speech 2 uses the CTC loss function to train a model on LibriSpeech. For inference, a transcription is generated using the trained language model.

% %\textbf{Training and Inference}:

% \textbf{LibriSpeech Dataset~\cite{panayotov2015librispeech}}: 
% % The dataset %is LibriSpeech~\cite{panayotov2015librispeech}, which is derived from audiobooks that are part of the LibriVox project, and 
% LibriSpeech is the most representative audio dataset and it contains 1000 hours of speech sampled at 16 kHz~\cite{panayotov2015librispeech}.

% \textbf{Reference Quality}: The word error rate (WER) of the reference implementation of DeepSpeech2 model on LibriSpeech validation data is 23.5\%.

% \subsubsection{Face Embedding}

\textbf{Face Embedding} is to verify a face by learning an embedding into the Euclidean space.
% and this can be used as face recognition, which is a critical area. 
%This task uses a 2D face dataset.% Verifying face typically uses CNN bottleneck layer or post-processing, which requires multiple models, PCA and SVM classification. Face embedding end-to-end trains the loss relevant to the face verification task and directly optimizes its performance.

% \textbf{FaceNet~\cite{schroff2015facenet}}: FaceNet model is a representative model and it is based on the GoogleNet style Inception model, which has about 24 million parameters. %Trade-offs are considered when implement the method and the best model may be different depending on application.

% \textbf{VGGFace2 Dataset~\cite{cao2018vggface2}}: % A large scale image dataset for face recognition. Images are downloaded from Google Image Search and 
% This dataset has large variations in pose, age, illumination, ethnicity, and profession, including 9000+ identities, and 3.3 million+ faces.

% \textbf{Reference Quality}: %In training stage, the network is robust with respect to JPEG quality and various embedding dimensionalities do not affect the performance significantly. For the amount of training data, the network has a good performance using tens of millions of exemplars and larger magnitude still gives small boost improvement in VAL. In inference stage, 
% The target quality is an accuracy of 90\%. %and get a classification accuracy of 95.12\%0.39 on Youtube Faces DB.

% \subsubsection{3D Face Recognition}

\textbf{3D Face Recognition} performs identification of 3D face images.
% which is an important task in developing, and has high requirements of reliability and stability. %this task uses a 3D face dataset which has different workload characteristics. %Compared to 2D counterpart, 3D face data carries more spatial information and is less sensitive to variations in poses, facial expressions and illuminations~\cite{zulqarnain2018learning}. 3D face recognition is becoming more beneficial for the facial similarity comparison and facial authentication scenario as 3D cameras become cheaper and more popular. There are usually 3 types of 3D image formats: depth image, point cloud, and mesh~\cite{zhou20183d}.

% \textbf{3D Face Model~\cite{vieriu2015facial}}: The model uses ResNet-50 network as backbone network and adjusts the first convolutional layer and the fully connect layer so that RGB-D images can be fed into the RGB-D ResNet-50 model. %The model is pretrained on 2D face datasets and then fine-tune it on the Intellifusion RGB-D datasets.

% \textbf{Intellifusion Dataset}: The dataset is an RGB-D dataset, provided by Intellifusion. %and crop images to 224x224 during preprocessing.% Each RGB-D image contains an ordinary RGB image and a depth image. %The value in the depth image reflects the distance of facial object surface from the viewpoint.

% \textbf{Reference Quality}: The reference implementation achieves an accuracy of 94.64\% on the Intellifusion dataset.

% \subsubsection{Recommendation}

%This task 
\textbf{Recommendation} is essential in the industry and widely used in commercial advertisements.

% \textbf{Neural collaborative filtering (CF)~\cite{he2017neural}}: CF is a fundamental algorithm for recommendation. Neural CF is a probabilistic approach using Gaussian assumptions on the known data and the factor matrices. %,  which can constrain priors to improve the algorithm, especially in the case of sparse rows.

% \textbf{MovieLens Dataset~\cite{harper2016movielens}}:% GroupLens Research has collected and made available rating data sets from the MovieLens web site (http://movielens.org). The data sets were collected over various periods of time, depending on the size of the set. 
% The MovieLens is a real-world movie ratings dataset from IMDB (the world's most popular and authoritative source for movie), The Movie DataBase, etc. The 100K movie ratings dataset contains 100,000 ratings from 1000 users on 1700 movies.

% \textbf{Reference Quality}: The quality metric is HR@10, which means whether the correct item is on the top-10 list. The target quality is 63.5\% HR@10.

% \subsubsection{Video Prediction}
\textbf{Video Prediction} 
% predicts how its actions affect objects in its environment, which is a representative video processing task. %And 
predicts the effect of physical interactions.
% which is a critical challenge for learning agents acting in the world, such as robots, autonomous cars, and drones.

% \textbf{Motion-Focused Predictive Model~\cite{finn2016unsupervised}}: This model predicts how to transform the last image into the next image. %As a result, it can reuse appearance information from previous frames and can better generalize to objects not seen in the training set.

% \textbf{Robot Pushing Dataset~\cite{finn2016unsupervised}}: This dataset contains 59,000 robot interactions involving pushing motions.% including a test set with novel objects. %In this dataset, accurate prediction of videos conditioned on the robot's future actions amounts to learning a ``visual imagination" of different futures based on different courses of action.

% \textbf{Reference Quality}: This task achieves 72 MSE on the test data.

% \subsubsection{Image Compression}
\textbf{Image Compression} aims to reduce the cost of storage or transmission using deep learning. 
% And bringing deep learning to image compression is an innovative work.% or transmitting digital images.

% \textbf{Recurrent Neural Network~\cite{toderici2017full}}: This model represents the RNN network, and it consists of a recurrent neural network (RNN)-based encoder and decoder, a binarizer, and a neural network for entropy coding. %The input images are first encoded into binary codes, then the decoder creates an estimate of the original input images based on the binary codes. The entropy coding layer improves the compression ratio.
% %During training, the residual error, the difference between the original image and the reconstruction from the decoder, is to be minimized.
% %During inference, only binary codes should be calculated. There are two training datasets.

% \textbf{ImageNet Dataset~\cite{deng2009imagenet}}: The dataset used for this task is the same as that of Image Classification.

% \textbf{Reference Quality}: The metric is 0.99 MS-SSIM (Multi-Scale-Structural Similarity Index~\cite{wang2003multiscale}).
% %The ï¬rst dataset is the one used in paper "Variable rate image compression with recurrent neural networks". The second dataset takes a random sample of 6 million 1280Ã720 images on the web. The model is evaluated on Kodak Photo CD dataset. Compared to previous work, models proposed in this paper shows improvements of 4.3\%â?.8\% AUC (area under the rate-distortion curve), depending on the perceptual metric used.

% \subsubsection{3D Object Reconstruction}
% %Understanding the 3D world is a fundamental problem in computer vision and computer graphics.
\textbf{3D Object Reconstruction} is to capture a real object's shape and appearance.
% a core technology of a wide variety of fields like computer graphics and virtual reality~\cite{3dreconstruction}. % using the most convenient 2D images.

% \textbf{Convolutional Encoder-decoder Network~\cite{yan2016perspective}}: This model combines image encoder, volume decoder, and perspective transformer.% leading to an end-to-end solution to 3D reconstruction.

% \textbf{ShapeNet Dataset~\cite{chang2015shapenet}}: ShapeNetCore contains about 51,300 unique 3D models from 55 common object
% categories. %Each 3D model is rendered from 24 azimuth angles with fixed elevation angles under the same camera and lighting setup. 

% \textbf{Reference Quality}: The metric is the average IU (intersection-over-union) score. %which is calculated over 24 volumes of all the instances in the test set. 
% The target average IU is 45.83\% on ShapeNetCore.

% \subsubsection{Text Summarization}
\textbf{Text Summarization} generates a headline or a summary and is an essential task in natural language processing. %like translation. %consisting of a few sentences that captures the salient ideas of an article or a passage. %which is important for search results preview, headline generation, and keyword discovery.

% \textbf{Sequence-to-sequence Model~\cite{nallapati2016abstractive}}: This model consists of an off-the-shelf attentional encoder-decoder RNN. % that was originally developed for machine translation to summarization. %For training, 200 dimensional word2vec vectors are trained on the same corpus to initialize the model embeddings, and allowed them to be updated during training.
% %For testing, full length F1 variant of Rouge is used to evaluate the model.

% \textbf{Gigaword Dataset~\cite{rush2017neural}}: The dataset contains about 3.8M training examples and 400K validation and test examples. %but we created a randomly sampled subset of 2000 examples each for validation and testing purposes.

% \textbf{Reference Quality}: The model achieves 41 Rouge-L on the Gigaword dataset.

% \subsubsection{Spatial Transformer}

\textbf{Spatial Transformer} provides spatial transformation capabilities, which we can use to process distorted and deformed objects.
% to improve other vision tasks' accuracy, such as image classification and object detection.

% \textbf{Spatial Transformer Network~\cite{jaderberg2015spatial}}: The model includes a localisation network, a grid generator, and a sampler. 
% %can actively learn a spatially transform matrix of an image (or a feature map). 
% %The transformation is performed on the entire feature map rather than locally. 
% %The transformation  includes scaling, cropping, rotations, as well as non-rigid deformations. 
% %The spatial transformer mechanism is split into three part a localisation network, a grid generator, a sampler.

% \textbf{MINST Dataset~\cite{lecun2010mnist}}: The MNIST dataset consists of  60,000 training images and 10,000 test images.
% %MNIST data has been distorted in various ways: rotation (R), rotation, scale and translation (RTS), projective transformation (P), and elastic warping (E). 

% \textbf{Reference Quality}: This task achieves an accuracy of 99\%.

% \subsubsection{Neural Architecture Search~\cite{elsken2018neural}}

\textbf{Neural Network Search} automatically designs neural networks.

\textbf{Advertising} is to display the most relevant ads to customers.

\textbf{Natural Language Processing (NLP)} is to train a language model, which we use for many tasks like translation and question answer.

\subsection{Reference Implementations and Metrics}~\label{rule_metric}

%This section briefly presents the reference implementations and evaluation metrics. 

\textbf{Reference Implementations.}
AIBench Training provides reference implementations and corresponding running scripts, datasets, and monitoring tools for each component benchmark. For each AI task, we provide implementations on both TensorFlow~\cite{abadi2016tensorflow} and PyTorch frameworks.

\textbf{Metrics.}
AIBench Training uses time-to-quality as a metric, which is the wall clock time in training a model to achieve a target quality~\cite{coleman2017dawnbench}. % focuses on a series of metrics covering accuracy, performance, and energy consumption, which are major concerns by our industry partners.

%\sout{The metrics for online inference contains query response latency, tail latency, throughput, inference accuracy, and inference energy consumption.}

%The metrics for AI training contains time-to-quality--the wall clock time to train a model achieving a target accuracy~\cite{coleman2017dawnbench}, the samples processed per second, the wall clock time to train the specific epochs, and the energy consumption to train a model achieving a target accuracy~\cite{coleman2017dawnbench}.

%% file: 4_evaluation.tex
\section{Evaluation}\label{evaluation}

This section %first conduct speedup analysis of AIBench component benchmarks on GPU against CPU (Section~\ref{speedupana}), and then 
compares AIBench Training (v1.1) against MLPerf Training (v0.7) (in short AIBench vs. MLPerf) from the perspectives of the model and micro-architectural characteristics (Section~\ref{7_bench_charac}), quantify the run-to-run variation, and measure their benchmarking cost (Section~\ref{7_cost}). Also, we propose the RPR and WC  subsets to achieve affordability and representativeness for repeatable performance ranking and workload characterization (Section~\ref{7_subset}), respectively. Further, we characterize micro-architectural behaviors from the perspectives of runtime breakdown, hotspot functions, and stall analysis (Section~\ref{7_micro}).

\subsection{Experimental Configuration}\label{Characterziation_Configuration}

We conducted experiments on two servers equipped with different GPUs: We use the TITAN XP GPUs for workload characterization and perform training experiments on the   TITAN RTX GPUs. The other configurations of the servers are the same, and the configurations of GPUs and servers are shown in Table~\ref{hwconfigeration}. The operating system is ubuntu 16.04 with the Linux kernel 4.4, and the other software is CUDA 10, python 3.7, and PyTorch 1.10. In the rest of this section, we mainly evaluate the reference PyTorch implementations of AIBench in Section~\ref{benchmark_cost}. The AIBench Traning homepage shows the performance numbers using other frameworks.

%We conducted experiments on two type servers equipped with different GPUs: one is TITAN XP, and the other is TITAN RTX. The other configurations of the servers are the same. The experiments for workload characterization are based on the TITAN XP GPUs, while the experiments running training sessions are based on TITAN RTX GPUs. The configurations of GPUs and other servers  are shown in Table~\ref{hwconfigeration}. The operating system we use is ubuntu 16.04 with the kernel version of Linux 4.4, and the other software is CUDA 10, python 3.7, and PyTorch 1.10. In the rest of this section, we only  evaluate the reference PyTorch implementations of AIBench because of the prohibitive training cost explained in Section~\ref{benchmark_cost}.

\begin{table}
  \caption{Hardware Configuration Details.}\label{hwconfigeration}
\renewcommand\arraystretch{1.1}
\center
\scriptsize
\begin{tabular}{|p{0.68in}|p{0.66in}|p{0.66in}|p{0.69in}|}
\hline \rowcolor{mygray} \multicolumn{4}{|l|}{CPU Configurations}\\
\hline \multicolumn{2}{|c|}{CPU Type} & \multicolumn{2}{c|}{Intel CPU Core} \\
\hline \multicolumn{2}{|c|}{Intel \textregistered Xeon E5-2620 v3}  &\multicolumn{2}{c|}{12 cores@2.40G} \\
\hline L1 DCache &L1 ICache &L2 Cache &L3 Cache \\
\hline 12 $\times$ 32 KB& 12 $\times$ 32 KB&12 $\times$ 256 KB& 15MB \\
\hline Memory & Ethernet & \multicolumn{2}{c|}{Hyper-Threading} \\
\hline 64GB, DDR3 & 1Gb & \multicolumn{2}{c|}{Disabled}  \\
%\hline \multicolumn{2}{|c|}{Memory} & \multicolumn{2}{c|}{32GB,DDR3}  \\
%\hline \multicolumn{2}{|c|}{Disk} & \multicolumn{2}{c|}{SATA@7200RPM}\\
%\hline \multicolumn{2}{|c|}{Ethernet} & \multicolumn{2}{c|}{1Gb}\\
%\hline \multicolumn{2}{|c|}{Memory} & \multicolumn{2}{c|}{64GB, DDR3}\\
%\hline \multicolumn{2}{|c|}{Ethernet} & \multicolumn{2}{c|}{1Gb}\\
%\hline \multicolumn{2}{|c|}{Hyper-Threading} & \multicolumn{2}{c|}{Disabled}\\
%\hline
%\hline \rowcolor{mygray} \multicolumn{4}{|l|}{Software Configurations}\\
%\hline \tabincell{l}{Operating\\System} & \tabincell{l}{Linux%\\Kernel} & \tabincell{l}{JDK\\Version} & \tabincell{l}{Hadoop\\Version} \\
%\hline CentOS 6.4 & 3.11.10 & 1.7.0 & 2.7.1\\
\hline \rowcolor{mygray} \multicolumn{4}{|l|}{GPU Configurations v1: Nvidia Titan XP}\\
%\hline \multicolumn{2}{|c|}{GPU Type} & \multicolumn{2}{c|}{Nvidia Titan XP} \\
\hline Cuda Cores & 3840 & Memory & 12GB, GDDR5X  \\
%\hline \multicolumn{2}{|c|}{Nvidia Cuda Cores} & \multicolumn{2}{c|}{3840 cores} \\
%\hline \multicolumn{2}{|c|}{GPU Memory} & \multicolumn{2}{c|}{12GB, GDDR5X} \\
\hline \rowcolor{mygray} \multicolumn{4}{|l|}{GPU Configurations v2: Nvidia Titan RTX}\\
%\hline \multicolumn{2}{|c|}{GPU Type} & \multicolumn{2}{c|}{Nvidia Titan RTX} \\
\hline Cuda Cores & 4608 & Memory & 24GB, GDDR6  \\
%\hline \multicolumn{2}{|c|}{Nvidia Cuda Cores} & \multicolumn{2}{c|}{4608 cores} \\
%\hline \multicolumn{2}{|c|}{GPU Memory} & \multicolumn{2}{c|}{24GB, GDDR6} \\
%\hline \rowcolor{mygray} \multicolumn{4}{|l|}{GPU Configurations v3: Nvidia Titan V}\\
%\hline \multicolumn{2}{|c|}{GPU Type} & \multicolumn{2}{c|}{Nvidia Titan V} \\
%\hline Cuda Cores & 5120 & Memory & 12GB, HBM2  \\
%\hline \multicolumn{2}{|c|}{Nvidia Cuda Cores} & \multicolumn{2}{c|}{5120 cores} \\
%\hline \multicolumn{2}{|c|}{GPU Memory} & \multicolumn{2}{c|}{12GB, HBM2} \\

\hline
\end{tabular}
\end{table}

\subsection{The Comparison of AIBench against MLPerf}\label{7_bench_charac}

\subsubsection{Model Characteristics}\label{attributes}

In this subsection, we first characterize the AIBench and MLPerf benchmarks' model characteristics from the perspectives of model complexity, computational cost, and convergent rate.

The characterization approach is like that of~\cite{bianco2018benchmark}. The differences have two points. First, they only evaluate different AI models of the same Image Classification task, and we assess twenty AI tasks and twenty-four models, which are the union set of AIBench and MLPerf. Second, they report Top-1 accuracy, and we report the convergent rate—the cost of training a model.  

We use the total amount of learnable parameters, FLOPs of a single forward computation, and the number of epochs to achieve a target quality (e.g., accuracy) to characterize the above three characteristics, respectively. We use the OpCounter~\cite{flopcounter} to estimate the FLOPs and learnable parameters for both AIBench and MLPerf benchmarks. Since the tool can not count some operations, the reported numbers may be smaller than the actual ones. We do not report the numbers of the reinforcement learning model for both AIBench (Neural Architecture Search) and MLPerf (Game) shown in Table~\ref{comparition_table} because the FLOPs and learnable parameters vary significantly from different epochs. We do not report the values of BERT and DLRM models for both AIBench (NLP, Advertising) and MLPerf (NLP, Recommendation) because the training is iteration-based, not epoch-based. The value of Video Prediction is not reported due to the limited support of OpCounter.

 For each benchmark of AIBench and MLPerf, we train the model to achieve a target quality. Specifically, the target quality is 74.9\% (accuracy) for Image Classification, 55\% (accuracy) for Text-to-Text translation, 4.2 (perplexity--the smaller is the better) for Image-to-Text, 23.5\% (WER--the smaller is the better) for Speech Recognition, 90\% (accuracy) for Face Embedding, 94.59\% (accuracy) for 3D Face Recognition, 76.7\% (mAP) for Object Detection, 60\% (HR@10) for Recommendation, 72 (MSE--the smaller is the better) for Video Prediction, 0.99 (MS-SSIM) for Image Compression, 45\% (IU) for 3D Object Reconstruction, 41 (Rouge-L) for Text Summarization, 99\% (accuracy) for Spatial Transformer, 13.9\% (accuracy) for Learning-to-Rank, 100 (perplexity) for Neural Architecture Search, 78.9\% for Advertising, and 71.2\% for Nature Language Processing.
 For the MLPerf benchmarks, the target quality is 37.7 (BBOX) for Object Detection (heavy), 22.47 (mAP) for Object Detection (light), 22.21 (BLEU) for Translation (recurrent), 25.25 (BLEU) for Translation (nonrecurrent). Note that AIBench and MLPerf use the same models and datasets for Image Classification, NLP, and Advertising (Recommendation task in MLPerf), so their numbers for these tasks are consistent in the rest of this paper.

\begin{figure}[tb]
\centering
\includegraphics[scale=0.40]{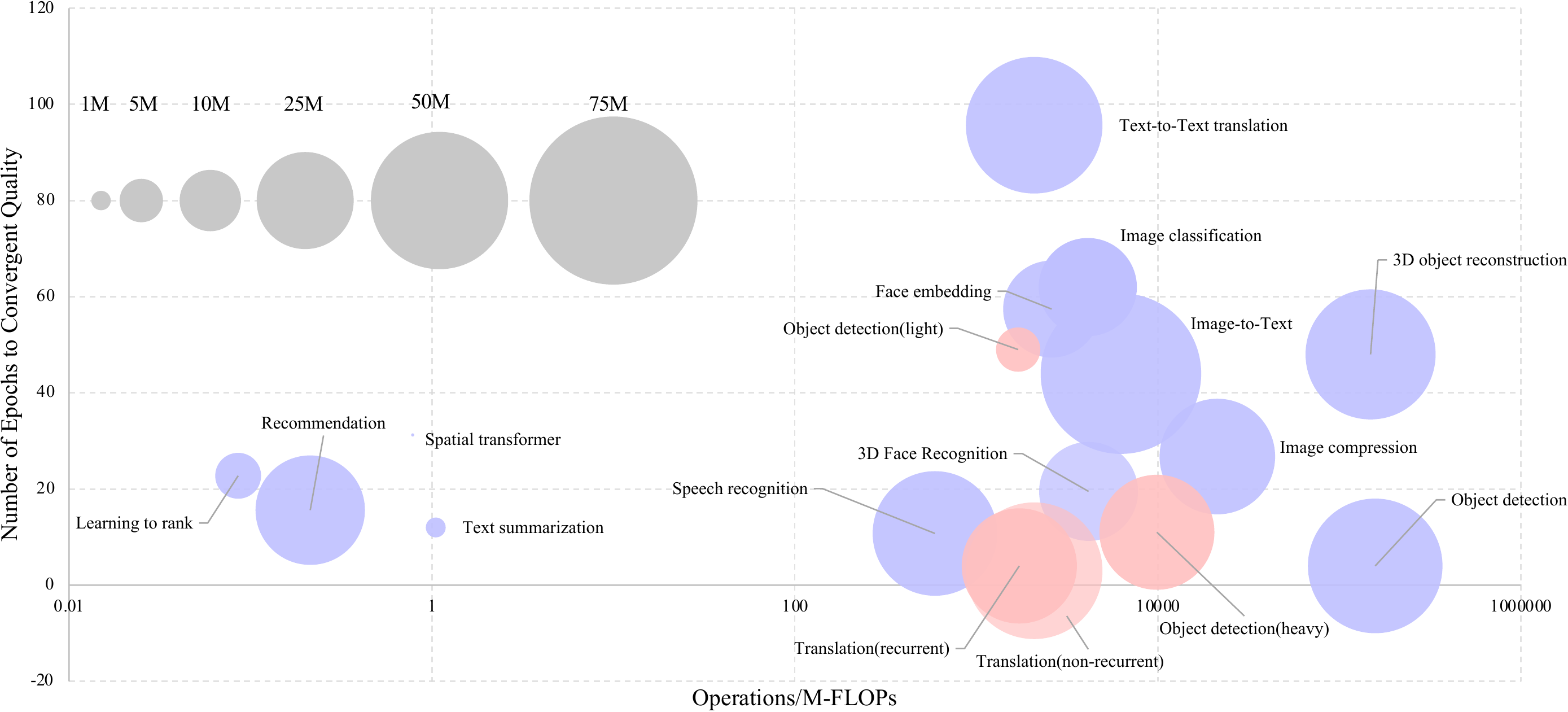}
\caption{The Comparisons of AIBench against MLPerf from the Perspectives of Model Complexity,  Computational Cost, and Convergent Rate.} %\vspace{1pt}
\label{workloads-characteristics}
\end{figure}

Fig.~\ref{workloads-characteristics} shows the model characteristics. From the computation cost perspective,  AIBench ranges from 0.09 to 282830 M-FLOPs, while MLPerf varies from 0.213248 to 24500 M-FLOPs--a much narrower range. From the perspective of model complexity, the number of learnable parameters of AIBench ranges from 0.03 million to 68.4 million, while MLPerf only covers a range of 5.2 to 49.53 million. From the convergent rate perspective, the required epochs of AIBench range from 6 to 96, while MLPerf only covers a range of 3 to 49. Thus, 
only using MLPerf cannot cover the diversities of different AI models.

Object Detection and 3D Object Reconstruction have the gigantic FLOPs, while Learning-to-Rank has the smallest. Image-to-Text is the most complex model, while the Spatial Transformer is the least.
% We find that Object Detection and 3D Object Reconstruction have the largest FLOPs among all benchmarks. Learning-to-Rank has the smallest number of FLOPs. Image-to-Text has the most complex model, while the Spatial Transformer has the least complex  model. %The full benchmarks of AIBench have the learnable parameters ranging  from 10 million to 100 million. 
%From the perspective of convergent rate, 
Text-to-text translation requires the largest epochs to converge, while the remaining models converge within 60 epochs.

Then we further investigate the optimizer and loss function categories. From the perspective of optimizers, both AIBench and MLPerf cover five optimizers, which are adam, adamw, RMSprop, SGD, and lamb for AIBench, and adam, lamb, lars, lazy, and SGD for MLPerf. From the perspective of loss function categories, AIBench covers fourteen loss functions, including BCELoss, BCEWithLogitsLoss, ChamferLoss, CrossEntropyLoss, CTCLoss, GANLoss, L1Loss, NLLLoss, SigmoidLogLoss, SmoothL1Loss, SoftmaxLoss, TripletLoss, Earth-Mover distance, and first order Euclidean distance, while MLPerf only covers six loss functions, which are BCELoss, CrossEntropyLos, LogLoss, SmoothL1Loss, SoftmaxLoss and VirtualLoss.

\subsubsection{Micro-architectural Characteristics}\label{access_patterns}

GPU architecture contains multiple streaming multiprocessors (SM), each of which has a certain number of CUDA cores, registers, caches, warp schedulers, etc. %Every CUDA core is an execute unit.
%Evaluating the GPU efficiencies against diverse AI benchmarks are significantly important for both the GPU architecture design and AI system optimization. In this section,
We choose five micro-architectural metrics to compare AIBench and MLPerf from a computation and memory access pattern perspective, including achieved occupancy, ipc\_efficiency, gld\_efficiency, gst\_efficiency, and dram\_utilization. Achieved\_occupancy represents the ratio of the average active warps per active cycle to the maximum number of warps provided by a multiprocessor~\cite{nvprof}. 
Ipc\_efficiency indicates the ratio of the executed instructions per cycle to the theoretical number~\cite{nvprof}.
Gld\_efficiency and gst\_efficiency represent the ratio of requested global memory load/store throughput to required global memory load/store throughput, respectively~\cite{nvprof}. Dram\_utilization means the utilization level of the device memory relative to peak utilization~\cite{nvprof}.
% Gld\_efficiency means the ratio of the requested global memory load throughput to the required global memory load throughput~\cite{nvprof}.
% Gst\_efficiency means the ratio of the requested global memory store throughput to the required global memory store throughput~\cite{nvprof}.
%single\_precision\_fu\_utilization represents the utilization level of the multiprocessor function units that execute single-precision floating-point instructions,
%tex\_utilization means the utilization level of the unified cache relative to the peak utilization,
%l2\_utilization means the utilization level of the L2 cache relative to the peak utilization,
%shared\_utilization means the utilization level of the shared memory relative to peak utilization,

\begin{figure}[tb]
\centering
\includegraphics[scale=0.65]{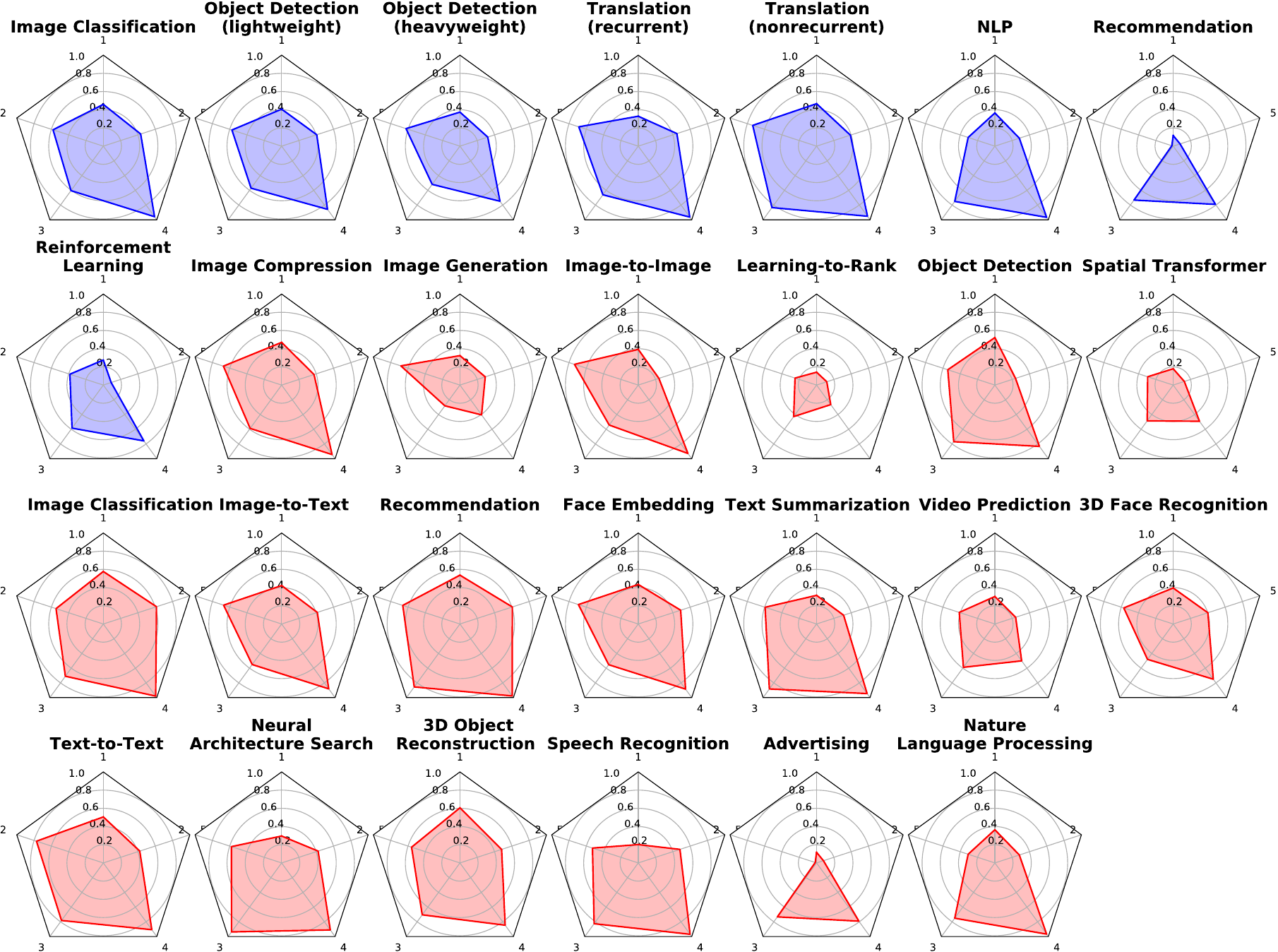}
\caption{The Computation and Memory Access Patterns of 24 Benchmarks from MLPerf (8) and AIBench (19)  (1: achieved\_occupancy; 2: ipc\_efficiency; 3: gld\_efficiency; 4: gst\_efficiency; 5: dram\_utilization). } %\vspace{1pt}
\label{mlperf_aibench_characteristics}
\end{figure}

Fig.~\ref{mlperf_aibench_characteristics} presents the computation and memory access patterns of the twenty-seven AI benchmarks (19 of AIBench, 8 of MLPerf). We find that AIBench captures distinct computation and memory patterns under different scenarios, e.g., processing text, image, audio, video, and various tasks of the same scenario, e.g., Image Classification and Image Generation, while MLPerf not.

Through Fig.~\ref{workloads-characteristics} and Fig.~\ref{mlperf_aibench_characteristics}, %Fig.~\ref{fig:subfig} shows 
we conclude that over AIBench, MLPerf has a significantly smaller coverage in terms of model complexity, computational cost, convergent rate, and computation and memory access patterns. AIBench outperforms MLPerf in terms of representativeness and diversity, which Table~\ref{comparition_table} also confirms.
%is a necessity. 
%diverse AI benchmarks reflecting different computation and memory access patterns should be included into the AI benchmarks. %Achieving  a state-of-the-art quality target for each AI task will incur heavy training overhead, however, it does not justify including only a few benchmarks~\cite{zhan2019benchcouncil}.

\subsection{Repeatability and Benchmarking Cost Evaluation}\label{7_cost}

%As the benchmarks should be repeatable, too large measurement errors justify not including some AI workloads as benchmarks. Meanwhile, the repeatability impacts the benchmarking cost significantly,  Because  an AI benchmark usually needs a set of runs to obtain the final result.
In this subsection, we describe the run-to-run variation and measure the benchmarking cost of AIBench against MLPerf. % against MLPerf.

\subsubsection{Run-to-run Variation}\label{variation}

Repeatability~\cite{bartlett2008reliability} refers to the variation in repeat measurements (different runs instead of different epochs using the same benchmark implementation under identical configurations) made on the same system under test.

Due to the stochastic nature of AI, different runs of the same benchmark under the same system are different. HPC AI500~\cite{jiang_hpc_2021} uses the coefficient of variation(the standard deviation / mean) of the number of epochs to quantify the run-to-run variation\footnote{For Advertising and Nature Language Processing, it uses the coefficient of variation of the number of iterations.}, and Table~\ref{AIBench_randomness_cost} (Column 5 and 6) shows the results.

\begin{table}[htbp]
\scriptsize
\caption{Training Costs and Run-to-run Variation (Column 5 and 6 are from HPC AI500~\cite{jiang_hpc_2021}) of Nineteen Benchmarks. The total time records the whole training time to achieve a target quality.}
\renewcommand\arraystretch{1.2}
\label{AIBench_randomness_cost}
\center %p{0.455in}|
\begin{tabular}{|p{0.3in}|p{1.5in}|p{0.45in}|p{0.39in}|p{0.39in}|p{0.39in}|}
\hline
\textbf{No.} & \textbf{Component Benchmark} & \textbf{Time Per Epoch (second)} & \textbf{Total Time (hour)} & \textbf{Variation} & \textbf{Repeat Times} \\
\hline
TrC1 & Image Classification & 4440 & 76.25 & 1.12\% & 8 \\
\hline
TrC2 & Image Generation & 3935.75 & N/A & N/A & N/A \\
\hline
TrC3 & Text-to-Text translation & 64.83 & 1.72 & 9.38\% & 6 \\
\hline
TrC4 & Image-to-Text & 845.02 & 10.21 & 23.53\% & 5 \\
\hline
TrC5 & Image-to-Image & 251.67 & N/A & N/A & N/A \\
\hline
TrC6 & Speech Recognition & 14326.86 & 42.78 & 12.08\% & 4 \\
\hline
TrC7 & Face Embedding & 214.73 & 3.43 & 5.73\% & 8 \\
\hline
TrC8 & 3D Face Recognition & 36.99 & 12.02 & 37.11\% & 10 \\
\hline
TrC9 & Object Detection & 1859.96 & 2.06 & 0 & 10 \\
\hline
TrC10 & Recommendation & 36.72 & 0.16 & 9.95\% & 5 \\
\hline
TrC11 & Video Prediction & 24.99 & 2.11 & 11.83\% & 4 \\
\hline
TrC12 & Image Compression & 763.44 & 5.67 & 22.49\% & 4 \\
\hline
TrC13 & 3D Object Reconstruction & 28.41 & 0.38 & 16.07\% & 4 \\
\hline
TrC14 & Text Summarization & 1923.33 & 6.41 & 24.72\% & 5 \\
\hline
TrC15 & Spatial Transformer & 6.38 & 0.06 & 7.29\% & 4 \\
\hline
TrC16 & Learning-to-Rank & 60.1 & 0.14 & 1.90\% & 4 \\
\hline
TrC17 & Neural Architecture Search & 932.79 & 7.47 & 6.15\% & 6 \\
\hline
TrC18 & Advertising & 1.77 & 2.28 & 0.12\% & 4 \\
\hline
TrC19 & Nature Language Processing (NLP)\tablefootnote{This benchmark uses the TensorFlow version, since the PyTorch version has no pre-trained model.} & 143.92 & 45.22 & 19.51\% & 5 \\
\hline
\end{tabular}
%\begin{tablenotes}
     %\item[1] 

  % \end{tablenotes}
\end{table}

\subsubsection{Evaluate Benchmarking Cost}\label{benchmark_cost}

%There are too many AI systems on the market, and each manufacturer advertises that its products are several times or even tens of times the performance of other products. It is an urgent matter to rank these AI systems fairly. 
%Section~\ref{7_bench_charac} describes the diversity and the necessity of all seventeen benchmarks for comprehensive benchmarking. However, 
Running entire training sessions of all AIBench benchmarks is prohibitively costly. Table~\ref{AIBench_randomness_cost} (Columns 3 and 4) lists the elapsed time for both an epoch and the total time for a training session to the target quality. 
% We find that 
Image Classification, Nature Language Processing, and Speech Recognition are the top-three most time-consuming benchmarks, which take about 164.25 hours in total. %Especially, because of the run-to-run variation mentioned in Section~\ref{variation}, an AI benchmark usually needs a set of runs to obtain the final result.
Supposing that we run each of the above three benchmarks five times, the time consumption reaches 34.2 days. If we do these for all nineteen benchmarks, the time will go 45.5 days, which is not affordable for most industries and academia. %Additionally, we find that 
%unbearable and unfeasible for ea. %So careful selection of representative workloads subsets is necessary. 
So, a concise, portable, and lightweight subset is of great significance.

Besides, we also evaluate the benchmarking cost of running a training session for MLPerf. To achieve a target quality, the time costs for MLPerf are 76.25 hours for Image Classification, 2.28 hours for Recommendation, 73.34 hours for Object Detection (heavyweight), 23.7 hours for Object Detection (lightweight), 16.52 hours for Translation (recurrent), 22 hours for Translation (nonrecurrent), 45.22 hours for NLP. For Reinforcement Learning %\footnote{Reinforcement learning still uses v0.5, which is similar to v0.7.} 
in MLPerf, we train the model for more than 96 hours, and the pro move prediction reaches 34\%, while the target is 40\%. Hence, running all seven benchmarks in MLPerf to achieve a target quality one time, the time cost goes more than 355.31 hours, even larger than running all benchmarks in AIBench. The reason is that for some workloads, AIBench uses a different model or dataset against MLPerf. For example, for Object Detection, AIBench uses Faster R-CNN, a fundamental model instead of Mask R-CNN, and uses the VOC dataset rather than the COCO dataset to achieve affordability. If we repeat MLPerf benchmarks five times, the time cost will be more than 74 days.

\subsection{The AIBench Subsets}\label{7_subset}

For performance ranking and workload characterization purposes, we choose two subsets.
%In this subsection, we illustrate how to choose these two subsets.% and the subsets' implications.

\subsubsection{How to Choose the Two Subsets?}\label{6_howsubset}

%However, ranking is different from system analysis and has several challenges.

We need to keep the subset repeatable, fair, affordable, and representative for the repeatable performance ranking purpose, so we keep the subset to a minimum from the following perspectives and criteria.
%as	few	as possible, to cover the above three requirements. As we consider the full benchmarks and the subset as two indispensable parts of the benchmark suite. %the model complexity, FLOPs, rate of convergence, run-to-run variation, time consumption to achieve a target quality, and evaluation metrics. 
%

\textbf{Reflecting diverse model complexity, computational cost, and convergent rate}. Specifically, we intend to choose the benchmarks that %exhibit representative speedup performance to avoid misleading or unfair benchmarking, and 
cover different aspects of Fig.~\ref{workloads-characteristics} as much as possible. 
% For example, the subset should cover a wide range of the number of FLOPs, learnable parameters, convergent rate.

\textbf{Run-to-run variation.} Repeatability is an important selection criterion of the subset. To avoid too much run-to-run variation, we choose the benchmarks with variance under 2\%.

\textbf{Widely accepted evaluation metrics.} A benchmark should have widely accepted performance metrics so that runs from different users have consistent termination conditions. So, we exclude the GAN-based models even if GANs are particularly important for content generation.

For the workload characterization purpose, we aim to select minimum workloads with the most representative system or micro-architectural characteristics, while the repeatability and consistent termination conditions are not mandatory.

% and will be considered for addition to the subset in the future if the issue of model quality evaluation can be resolved.

%The first challenge is affordability. There are too many workloads in AIBench, and it will be time-consuming if all the tests are done. 

%The second challenge is the repeatability of the benchmark due to the randomness of AI training. Although for some loads, the training process can eliminate randomness under the condition of fixing all hyperparameters and random seeds, this will cause the AI system not fully to present its own performance. Therefore, for ranking AI systems, the load with relatively low randomness should be selected as much as possible.

%The last challenge is when to terminate AI training. We will face two problems. One is that some workloads do not currently have clear evaluation metrics, such as the GAN model. Trainers need to rely on the eye to judge whether the model is well trained, which is very subjective. Another problem is to choose a suitable accuracy threshold. If the accuracy threshold is set too high, it may cause some AI systems never to train the model to the specified accuracy, and different accuracy may result in different randomness of the model. Therefore, it is essential to select a workload with a precise evaluation metric and reasonably determine the threshold of accuracy.

\subsubsection{The Subsets Decision}\label{6_subset}

% We include three benchmarks into the benchmarking subset:
AIBench Training 
provides two subsets for repeatable performance ranking (RPR subset) and workload characterization (WC subset) to improve affordability. %, according to the above criteria illustrated in Section~\ref{6_howsubset}, we choose RPR subset and WC subset.

\textbf{RPR subset}. The RPR subset includes three benchmarks: Image Classification, Object Detection, and Learning-to-Rank. To satisfy the first criterion, they cover different ranges of numbers in terms of FLOPs and learnable parameters (both small for Learning-to-Rank, medium for Image Classification, and large for Object Detection), and different convergent rates (small epochs for Object Detection, medium for Learning-to-Rank, and large for Image Classification).
As for the second criterion, three benchmarks have the low run-to-run variation, 1.12\% for Image Classification, 1.9\% for Learning-to-Rank, and 0\% for Object Detection. Besides, they have widely accepted evaluation metrics.

\textbf{WC subset}. The WC subset also includes three benchmarks: Spatial Transformer, Image-to-Text, and Speech-to-Text. This subset reflects the most representative micro-architectural workload characteristics since they are the nearest to the centroid of three clusters, respectively, according to the K-means clustering results on AIBench Training benchmarks in~\cite{jiang_hpc_2021}.

Comparing to the full benchmark of AIBench and MLPerf, the AIBench RPR subset shortens the training time by 64\% and 78\%, respectively, and the AIBench WC subset shortens by 76\% and 85\%, respectively.

\subsection{Micro-architectural Behaviors}\label{7_micro}

This subsection characterizes the micro-architectural behaviors from the perspectives of runtime breakdown, hotspot function analysis, and stall analysis.

\subsubsection{Runtime Breakdown}\label{microarch}

 %We characterize the computation and memory access patterns of the seventeen AI tasks, to further emphasize the necessity of including diverse AI tasks for benchmarking. 
 
%We characterize the seventeen component benchmarks of AIBench.
 %GPU architecture contains multiple streaming multiprocessors (SM), each of which has a certain number of CUDA cores, memory registers, memory caches, warp schedulers and etc. 
% Evaluating the GPU efficiencies against diverse AI benchmarks are significantly important for both the GPU architecture design and AI system optimization. Hence, we deploy the AIBench component benchmarks on the Titan XP GPUs, and focus on a single GPU performance. The CUDA and Nvidia driver versions are 10.0 and 410.78, respectively. 
%on the Titan XP GPU and explore their GPU execution efficiencies.  
%We choose the PyTorch implementations with the version of 1.1.0 for evaluation.
%Further, we perform runtime breakdown of these component benchmarks.
We evaluate the PyTorch implementations with Pytorch 1.1.0. The dataset for each benchmark is as follows: ImageNet (137 GB) for Image Classification and Image Compression; LSUN (42.8 GB) for Image Generation; VGGFace2 (36 GB) for Face Embedding; Microsoft COCO (13 GB) for Image-to-Text; VOC2007 (439 MB) for Object Detection; MNIST (9.5 MB) for Spatial Transformer; Cityscapes (267 MB) for Image-to-Image; MovieLens (190 MB) for Recommendation; Librispeech (59.3 GB) for Speech Recognition; Gowalla (107 MB) for Learning-to-Rank; WMT English-German (1.2 MB) for Text-to-Text translation; Robot pushing dataset (137 GB) for Video Prediction; ShapeNet dataset (6.8 GB) for 3D Object Reconstruction; Gigaword dataset (277 MB) for Text Summarization; 3D face data (37 GB) for 3D Face Recognition; PTB dataset (4.9 MB) for Neural Architecture Search, respectively, Kaggle Display Advertising Challenge Dataset (4.3 GB) for Advertising, and Wikipedia (76 GB) for NLP.

 The overall execution performance of these component benchmarks varies in terms of IPC, which measures the executed instructions per cycle. Fig.~\ref{mlperf_aibench_characteristics} shows that the IPC efficiency ranges from 0.02 (Advertising) to 0.77  (Text-to-Text translation). Some benchmarks like Learning-to-Rank have extremely low IPC comparing to other benchmarks. To reveal the factors that impact the performance significantly, we first conduct runtime breakdown analysis and decompose the benchmarks into the hotspot functions. Then we explore the GPU execution efficiency in terms of different percentages of stalls.

\begin{figure}[tb]
\centering
\includegraphics[scale=0.66]{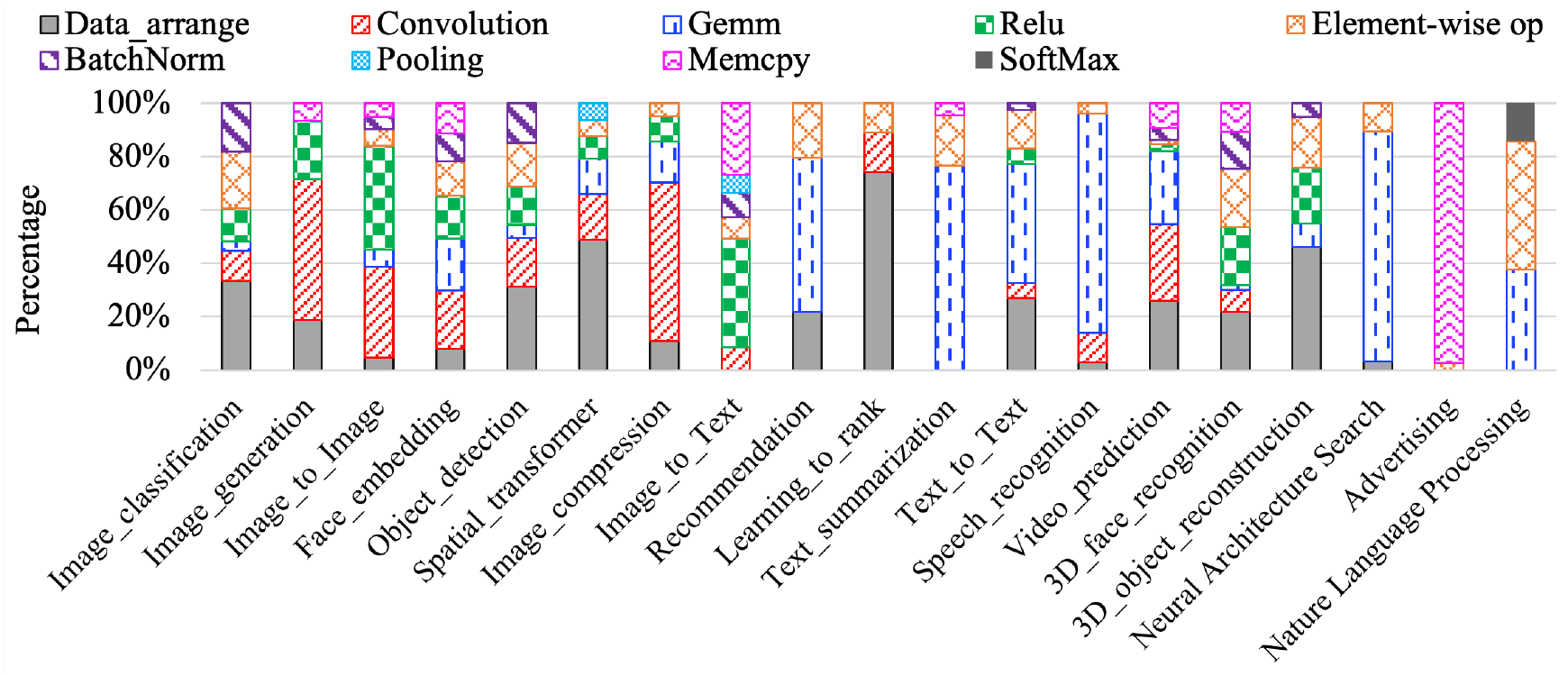}
\caption{Runtime Breakdown of the AIBench Benchmarks.} %\vspace{1pt}
\label{time-breakdown}
\end{figure}

We use NVProf to trace the runtime breakdown and find the hotspot functions that occupy more than 80\% of runtime in total.
Since each run involves dozens of function calls, we single out the functions that occupy large proportions of runtime and classify them into several categories of kernels according to their computation logic.
Through statistics, we find that the most time-consuming functions among all component benchmarks have much in common, and they are classified into nine categories of kernels: data arrangement, convolution, general matrix multiply (gemm), batch normalization, element-wise operation, relu activation, %~\footnote{Relu activation is an element-wise operation, here we use a separate category of Relu considering its large proportion and diverse CUDA functions.}
pooling, memory copy, and softmax, spanning from computation kernels to memory access kernels.
%, which are consistent with our micro benchmarks and further indicates the decision of including them is correct. 
Note that each kernel contains a bunch of functions that solve a similar issue. For example, a gemm kernel includes single or double precision floating general matrix multiply.
Fig.~\ref{time-breakdown} shows the runtime breakdown of nineteen benchmarks of AIBench, using the average number of all involved functions within each category of kernels. Note that this figure does not consider the remaining 20\% functions.
Further, we summarize typical functions for each category of kernels that occupy a large proportion of runtime among the component benchmarks, as shown in Table~\ref{func-sum}. We find that Learning-to-Rank spends too much time on data arrangement operations from Fig.~\ref{time-breakdown}, and the corresponding function call is maxwell\_scudnn\_128x32\_stridedB\_splitK\_interior\_nn with an IPC of 0.98. This is the reason that Learning-to-Rank has the low IPC of 0.99.
We believe that the nine categories of kernels and these corresponding functions are the optimization points for CUDA library optimizations and micro-architectural optimizations.

\begin{table}[htbp]
\scriptsize
\caption{Hotspot Functions.}
\renewcommand\arraystretch{1.2}
\scriptsize
\label{func-sum}
\center %p{0.455in}|
\begin{tabular}{|p{0.76in}|p{3.45in}|}
\hline
\textbf{Micro Benchmark} & \textbf{Function Name} \\
\hline
%\multirow{3}{*}{
Data Arrangement & \tabincell{l}{1) maxwell\_scudnn\_128x128\_stridedB\_splitK\_interior\_nn; \\2) maxwell\_scudnn\_128x32\_stridedB\_splitK\_interior\_nn;\\ 3) maxwell\_scudnn\_128x128\_stridedB\_interior\_nn} \\
%\cline{2-2}
%& maxwell\_scudnn\_128x32\_stridedB\_splitK\_interior\_nn\\
%\cline{2-2}
%& maxwell\_scudnn\_128x128\_stridedB\_interior\_nn \\
\hline
Convolution &  \tabincell{l}{1) maxwell\_scudnn\_winograd\_128x128\_ldg1\_ldg4\_tile148n\_nt; \\2) wgrad\_alg0\_engine; 3) fft2d\_r2c\_32x32}\\
%\cline{2-2}
%& wgrad\_alg0\_engine \\
%\cline{2-2}
% & fft2d\_r2c\_32x32 \\
\hline
GEMM & \tabincell{l}{1) maxwell\_sgemm\_128x64\_nt;\\ 2) maxwell\_sgemm\_128x64\_nn; \\3) sgemm\_32x32x32\_NN\_vec;} \\
%\cline{2-2}
%& maxwell\_sgemm\_128x64\_nn \\
%\cline{2-2}
%& sgemm\_32x32x32\_NN\_vec \\
\hline
BatchNorm &  \tabincell{l}{1) cudnn::detail::bn\_fw\_tr\_1C11\_kernel\_NCHW; \\2) cudnn::detail::bn\_bw\_1C11\_kernel\_new; \\3) batch\_norm\_backward\_kernel; \\4) at::native::batch\_norm\_backward\_kernel}  \\
%\cline{2-2}
%& cudnn::detail::bn\_bw\_1C11\_kernel\_new  \\
%\cline{2-2}
%& batch\_norm\_backward\_kernel\\
%\cline{2-2}
%& at::native::batch\_norm\_backward\_kernel \\
\hline
Relu & \tabincell{l}{1) maxwell\_scudnn\_128x128\_relu\_small\_nn; \\2) maxwell\_scudnn\_128x128\_relu\_interior\_nn;\\ 3) maxwell\_scudnn\_128x32\_relu\_interior\_nn } \\
%\cline{2-2}
%& maxwell\_scudnn\_128x128\_relu\_interior\_nn  \\
%\cline{2-2}
%& maxwell\_scudnn\_128x32\_relu\_interior\_nn \\
\hline
Element-wise & \tabincell{l}{1) element-wise add kernel; 2) element-wise threshold kernel; \\3) element-wise mul kernel } \\
%\cline{2-2}
%& element-wise threshold kernel \\
%\cline{2-2}
%& element-wise mul kernel \\
\hline
\multirow{1}{*}{Pooling} & 1) MaxPoolBackward; 2) AvePoolForward \\
%\cline{2-2}
%& AvePoolForward \\
\hline
\multirow{1}{*}{Memcpy} & 1) CUDA memcpy HtoD; 2) CUDA memcpy DtoD \\
%\cline{2-2}
%& CUDA memcpy DtoD \\
\hline
\multirow{1}{*}{Softmax} & 1) SoftMax\_compute \\
\hline

\end{tabular}
\end{table}

\subsubsection{Hotspot Function Analysis}

Hotspot function identification is of great significance for bottleneck locating and code optimization. We compare the essential hotspot functions identified by AIBench and MLPerf.

\begin{figure}[tb]
\centering
\includegraphics[scale=0.75]{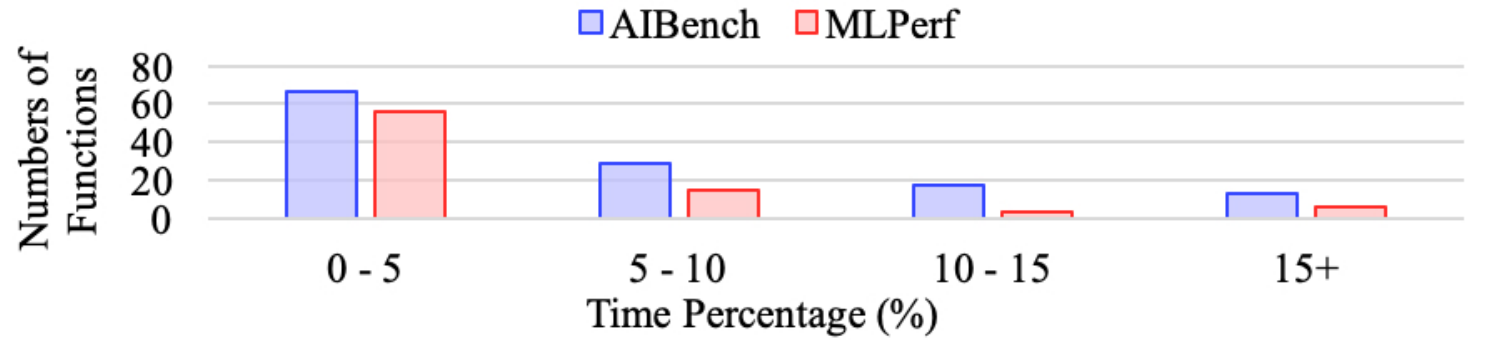}
\caption{The Number of Hotspot Functions Identified by AIBench and MLPerf.} %\vspace{1pt}
\label{number_of_functions_in_different_time_range}
\end{figure}

Fig.~\ref{number_of_functions_in_different_time_range} shows the numbers of hotspot functions within each category of occupying different time percentages identified by AIBench and MLPerf. 
%lists the major hotspot functions of seventeen component benchmarks and MLPerf benchmarks, respectively. 
We find that MLPerf only covers a fraction of hotspot functions over AIBench. For example, within the category that occupies more than 10\% of runtime, the number of hotspot functions profiled from AIBench is 30, while only 9 for MLPerf. Thus, MLPerf omits many hotspot functions that occurred in a broad spectrum of AI tasks. %On the contrary, AIBench has the ability to find more hotspot functions in important AI tasks. 

We further profile the AIBench RPR subset to find whether they capture the primary hotspot functions. Our evaluation shows that even though the RPR subset captures the least number of hotspot functions compared to the full benchmarks of AIBench and MLPerf, it still covers the most time-consuming and frequently-appearing functions within these benchmarks like  maxwell\_scudnn\_128x128\_stridedB\_splitK\_interior\_nn (e.g., occupying 17\% running time for 3D Object Reconstruction). %In addition, from the perspective of benchmarking cost, comparing to seventeen benchmarks and MLPerf, the subset shortens the training time by \textcolor[rgb]{1,0,0}{xx} and \textcolor[rgb]{1,0,0}{xx} times, respectively. 
 
In conclusion, for AIBench, the full benchmarks and the subset are two indispensable parts. Over MLPerf, the complete benchmarks of AIBench provide comprehensive workload characterization and detailed evaluation while reducing the training time by 37\%. Against the AIbench full benchmarks, its subset further shortens the benchmarking cost by 41\% while maintaining the primary workload characteristics.
%, and hence is beneficial for entire training and performance ranking.

\subsubsection{Stall Analysis}

We analyze the percentage of stalls of eight kinds of stalls focusing on the above eight kernel categories. Instruction fetch stall (Inst\_fetch) indicates the next assembly instruction has not yet been fetched. Execution dependency stall (Exe\_depend) is because an input required by the instruction is not, however, available. Memory dependency stall (Mem\_depend) is because a memory operation cannot be performed due to the required resources not being available or fully utilized. Texture stall (Texture) is because of the under-utilization of the texture sub-system.  Synchronization stall (Sync) is due to a syncthreads call. Constant memory dependency stall (Const\_mem\_depend) is because of an immediate constant cache miss. Pipe busy stall (Pipi\_busy) is because a compute operation cannot be performed because the compute pipeline is busy. Memory throttle stall (Mem\_throttle) is due to large pending memory operations~\cite{nvprof}.

\begin{figure}[tb]
\centering
\includegraphics[scale=0.85]{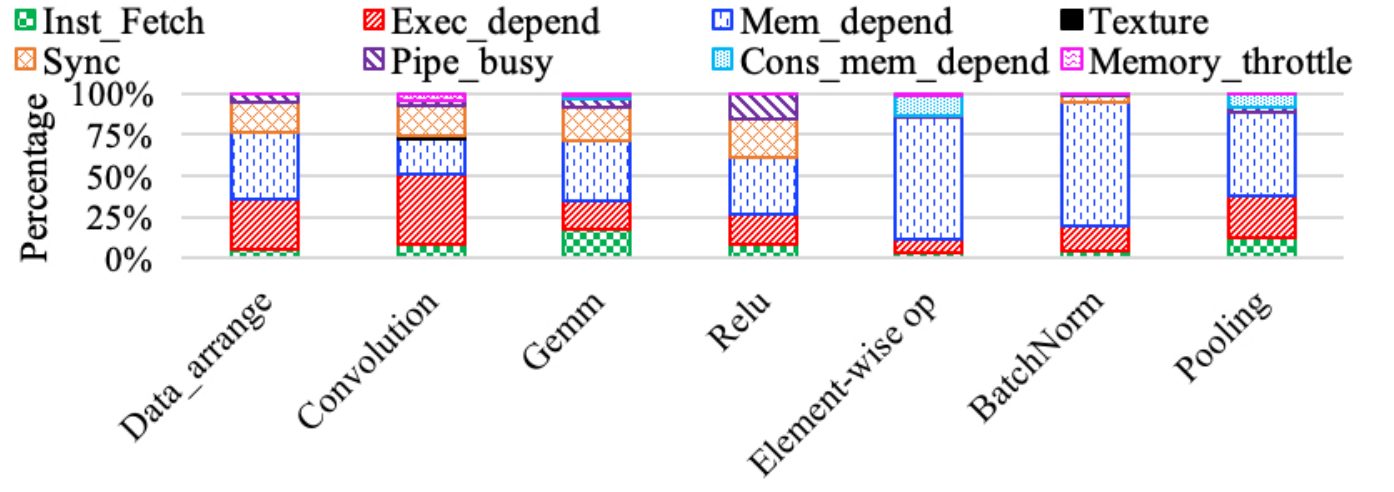}
\caption{Stall Breakdown of the Hotspot Functions.} %\vspace{1pt}
\label{stall-breakdown}
\end{figure}

%In order to find the micro-architectural bottleneck, we further do the stall analysis at the functional level.
 Fig.~\ref{stall-breakdown} shows the breakdown of eight stalls of the hotspot functions.
The top two GPU execution stalls are memory dependency stalls and execution dependency stalls. For example, for Element-Wise kernels, the memory dependency stalls occupy a substantial proportion of 70\%, resulting in a low IPC of 0.86 on average.
The memory dependency stalls may occur due to high cache misses.%, and thus the load/store resources are not available.
Possible optimizations include optimizing data alignment, data locality, and data access patterns. The execution dependency stalls may occur due to low instruction-level parallelism, and exploiting ILP may alleviate partial execution dependency stalls.

%% file: 5_performance_ranking.tex
\section{Performance Ranking}~\label{Section_ranking}

In this section, we use the AIBench RPR subset to rank and report the performance of GPUs and TPUs.
% following the system-level benchmarking rules. 
For GPU evaluations, we deploy them on our local servers. For TPU evaluation, we rent a Google Cloud environment. Table~\ref{ranking-fig} lists their configurations. Note that the other server configurations 
for GPUs are consistent with the illustrations in Section~\ref{Characterziation_Configuration}.

\begin{table}[htbp]
\scriptsize
\caption{GPU and TPU Configurations.}
\label{ranking-fig}
\center %p{0.455in}|
\begin{tabular}{|p{0.53in}|p{0.28in}|p{0.48in}|p{0.35in}|p{0.6in}|p{0.21in}|}
\hline
%\rowcolor{mygray} \multicolumn{6}{|l|}{GPU Configurations}\\
%\hline
\rowcolor{mygray} \textbf{GPU Type} & \textbf{Arch} & \textbf{GPU Cores} & \textbf{Memory} & \textbf{FP32} & \textbf{Year} \\
\hline
RTX 2080S &	Turing&	3072&	8GB&	11.15 TFLOPS&	2019 \\
\hline
RTX 2080 Ti & Turing & 4352 & 11GB & 13.1 TFLOPS & 2018 \\
\hline
TITAN RTX & Turing & 4608 & 24GB & 16.3 TFLOPS & 2018\\
\hline
TITAN XP	&	Pascal&	3840&		12GB&	12.15 TFLOPS&2017\\
\hline
TITAN V & Volta & 5120 & 12GB & 15 TFLOPS & 2017\\
\hline
P100 & Pascal & 3584 & 16GB & 9.3 TFLOPS &2016 \\
\hline
P40 & Pascal & 3840 & 24GB & 12 TFLOPS & 2016\\
\hline
%\rowcolor{mygray} \multicolumn{6}{|l|}{TPU Configurations}\\
%\hline
\rowcolor{mygray} \textbf{TPU Type} & \textbf{Arch} & \textbf{TPU Cores} & \textbf{Memory} & \textbf{Bfloat16} & \textbf{Year} \\
\hline
TPU v3 & N/A & 8 & 128GB & 420 TFLOPS & 2018 \\
\hline
TPU v2 & N/A & 8 & 64GB & 180 TFLOPS & 2017 \\
\hline
\end{tabular}
\end{table}

%\subsection{Single-Card Training Ranking}

\begin{figure}[tb]
\centering
\includegraphics[scale=0.95]{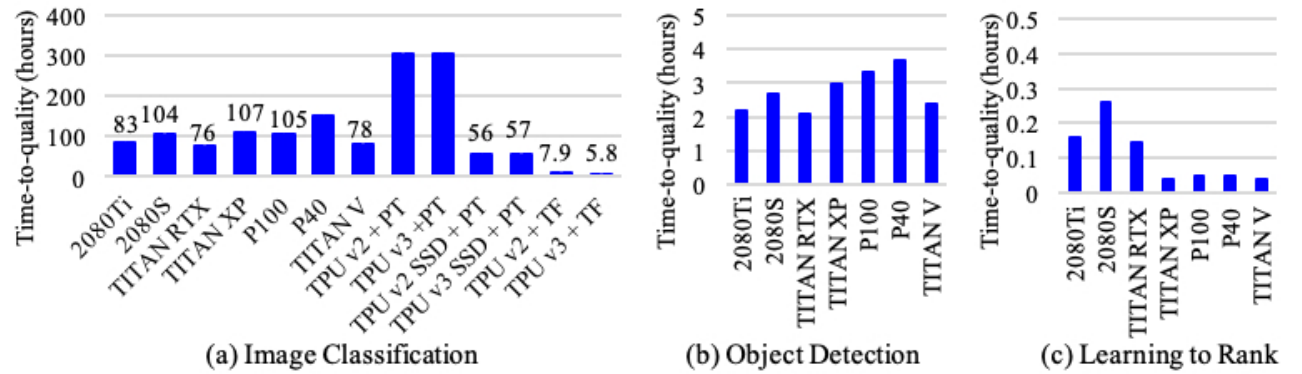}
\caption{GPU and TPU Ranking using AIBench RPR subset. PT and TF indicate PyTorch and TensorFlow versions. SSD represents a solid-state disk. TPU currently does not support Object Detection (Faster R-CNN) and Learning-to-Rank~\cite{tpusupport}.} %\vspace{1pt}
\label{ranking-result}
\end{figure}

Our benchmarks support distributed training, and the performance difference of multi-cards is consistent with single-card, so we only report single-card performance ranking here. 
% We evaluate the single-card performance ranking of 
We evaluate the GPU and TPU listed in Table~\ref{ranking-fig} and use the PyTorch version of the AIBench RPR subset. The metric is time-to-quality. As shown in Fig.~\ref{ranking-result}, the RPR subset has deficient performance on TPU compared to GPU, with a deterioration of 2 to 4 times. The reason is that
TPU connects the host virtual machine of Google Cloud through the network instead of being embedded on the motherboard like GPU. Hence, the data feeding is highly input-bound for Image Classification unless there are many workers to feed in data and sufficient RAM to maintain many worker threads~\cite{tpuforpytorch}. Our experiments show that when using TPU, the average I/O Wait ratio of the virtual machine is more than 80\%, which is indeed a severe I/O bottleneck.
%although TPU provides a mechanism to read the data in a distributed way from cloud storage directly and cache them into memory to achieve  fast data access, however, PyTorch cannot use this mechanism and have to read data from the virtual machine, and further results in severe I/O bottleneck. Our experiments show that when using TPU, the average I/O Wait ratio of the virtual machine is more than 80\%.

Considering the I/O bottleneck of PyTorch on TPU, we replace HDD with SSD to accelerate I/O processing (TPU v2/3 SSD+PT in Fig.~\ref{ranking-result}(a)). Image classification performance improves about 5.3 times compared to the original for both TPU v2 and v3 and is even better than all tested GPUs. We further evaluate the TensorFlow version's performance on TPU (TPU v2/3+TF in  Fig.~\ref{ranking-result}(a)). The TensorFlow version on TPU achieves the best performance. For Image Classification, it spends 7.8 hours on TPU v2 and 5.8 hours on TPU v3, which is about eight times higher than that of the PyTorch version on TPU with SSD average, and 9.6 times higher than the best performance on GPU (TITAN RTX: 76 hours). The reason is that TPU provides a mechanism specialized for TensorFlow, 
which reads the data in a distributed way from cloud storage directly and caches them into memory to fully utilize the high-speed network bandwidth and achieve fast data access. Although TPU reflects extremely high performance for Image Classification, it supports limited models officially, bringing the huge portability cost, which is not the case for many general-purpose accelerators like GPUs.
%\subsection{Multi-card Training Ranking}
%
%Due to the quantitative limitation of TPU, we evaluate the multi-card training performance on GPUs.

%% file: 6_conclusion.tex
\section{Conclusion}\label{section_conclusion}
% 詹老师负责

This paper summarizes AI benchmarking challenges: prohibitive cost, conflicting requirements,  short shelf-life, scalability, and repeatability. AIBench is the first benchmark project that systematically tackles the above challenges: it distills and abstracts real-world application scenarios into the scenario, training, inference, micro, and synthetic AI benchmarks. 

We present a balanced industry-standard methodology to meet the conflicting benchmarking requirements in different stages.
We use real-world benchmarks to cover the factors space that impacts the learning dynamics to the most considerable extent. %This methodology can guarantee benchmarks' diversity and representativeness.  
We identify and implement nineteen representative AI tasks with state-of-the-art models.
For two different purposes: repeatable performance ranking (RPR subset) and workload characterization (WC subset), we keep two subsets to a minimum for affordability.
%We provide the microbenchmarks for simulation-based architecture research. 
%We  contribute by far the most comprehensive AI training benchmark suite.

%This paper presents a balanced AI benchmarking methodology. %that  meets with the conflicting requirements in different stages of industry-standard AI benchmarking. 
%We present a comprehensive component AI benchmark suite—AIBench and its subset as two indispensable parts.
% AIBench Subset is affordable and can be used for evaluating new architectures or systems in early-stage, and the full benchmark suite can be used for detailed evaluation in later-stage.
%, and contribute by far the most comprehensive open-source industry-standard AI benchmark suite---AIBench.
 %We include as most as possible (nineteen) representative AI tasks from one of the most important domains—Internet Services, to guarantee the representativeness and diversity of the benchmarks. Meanwhile, we keep a subset to a minimum (three tasks) for affordability.  
  %We , and run   the  subset and/or selectively run  the full benchmarks to reduce the benchmarking cost while avoid error-prone design  or benchmarketing. 
 % We perform by far the most comprehensive workload characterization on AIBench and MLPerf.
  %Training from the perspectives of model complexity, computational cost, and  convergent rate,  computation and memory access patterns, repeatability, hotspot functions, and other micro-architecture characteristics. 
  Our evaluations show AIBench outperforms MLPerf in terms of the diversity and representativeness of model complexity, computational cost, convergent rate, computation and memory access patterns, and hotspot functions; %in terms of the diversity and representativeness of model complexity, computational cost, convergent rate, computation and memory access patterns, and hotspot functions.
  %With respect to its counterpart, 
  AIBench reduces the benchmarking cost while avoiding error-prone design or benchmarketing. With respect to the AIBench full benchmarks, its RPR subset shortens the benchmarking cost by 64\%, while maintaining the primary workload characteristics.